\documentclass[10pt,twocolumn,letterpaper]{article}

\usepackage[pagenumbers]{cvpr} % To force page numbers, e.g. for an arXiv version
\usepackage{graphicx}
\usepackage{amsmath}
\usepackage{amssymb}
\usepackage{booktabs}

\usepackage{mathtools}
\usepackage{microtype}
\usepackage{subcaption}
\usepackage{cite}
\usepackage{bm}
\usepackage{multirow}
\usepackage{float}
\usepackage{array}
\usepackage{duckuments}
\newcolumntype{P}[1]{>{\centering\arraybackslash}m{#1}}
\DeclareMathOperator*{\argmax}{arg\,max}

\usepackage[pagebackref,breaklinks,colorlinks]{hyperref}

\usepackage[capitalize]{cleveref}
\crefname{section}{Sec.}{Secs.}
\Crefname{section}{Section}{Sections}
\Crefname{table}{Table}{Tables}
\crefname{table}{Tab.}{Tabs.}

\usepackage{pifont}% http://ctan.org/pkg/pifont
\newcommand{\cmark}{\ding{51}}%
\newcommand{\xmark}{\ding{55}}%

\def\cvprPaperID{11803} % *** Enter the CVPR Paper ID here
\def\confName{CVPR}
\def\confYear{2023}

\usepackage{lipsum}

\begin{document}

\title{IFSeg: Image-free Semantic Segmentation via Vision-Language Model}

\author{Sukmin Yun$^{13}\thanks{Equal contribution}\;\thanks{Work was done while at KAIST}$\quad\quad Seong Hyeon Park$^1$\footnotemark[1]\quad\quad Paul Hongsuck Seo$^2$\quad\quad Jinwoo Shin$^1$\\
$^1$Korea Advanced Institute of Science and Technology (KAIST) \quad $^2$Google Research\\
$^3$Mohamed bin Zayed University of Artificial Intelligence (MBZUAI)\\
{\tt\small sukmin.yun@mbzuai.ac.ae, seonghyp@kaist.ac.kr, phseo@google.com, jinwoos@kaist.ac.kr}\\
}
\maketitle

%%%%%%%%% ABSTRACT
\begin{abstract}
Vision-language (VL) pre-training has recently gained much attention for its transferability and flexibility in novel concepts (e.g., cross-modality transfer) across various visual tasks.
However, VL-driven segmentation has been under-explored, and the existing approaches still have the burden of acquiring additional training images or even segmentation annotations to adapt a VL model to downstream segmentation tasks.
In this paper, we introduce a novel image-free segmentation task where the goal is to perform semantic segmentation given only a set of the target semantic categories, but without any task-specific images and annotations.
To tackle this challenging task, our proposed method, coined IFSeg, generates VL-driven artificial image-segmentation pairs and updates a pre-trained VL model to a segmentation task. 
We construct this artificial training data by creating a 2D map of random semantic categories and another map of their corresponding word tokens.
Given that a pre-trained VL model projects visual and text tokens into a common space where tokens that share the semantics are located closely, this artificially generated word map can replace the real image inputs
for such a VL model.
Through an extensive set of experiments, our model not only establishes an effective baseline for this novel task but also demonstrates strong performances compared to existing methods that rely on stronger supervision, such as task-specific images and segmentation masks. Code is available at \url{https://github.com/alinlab/ifseg}.
\end{abstract}

%%%%%%%%% BODY TEXT
\section{Introduction}
\begin{figure}[t]
  \centering
   \includegraphics[width=0.95\linewidth]{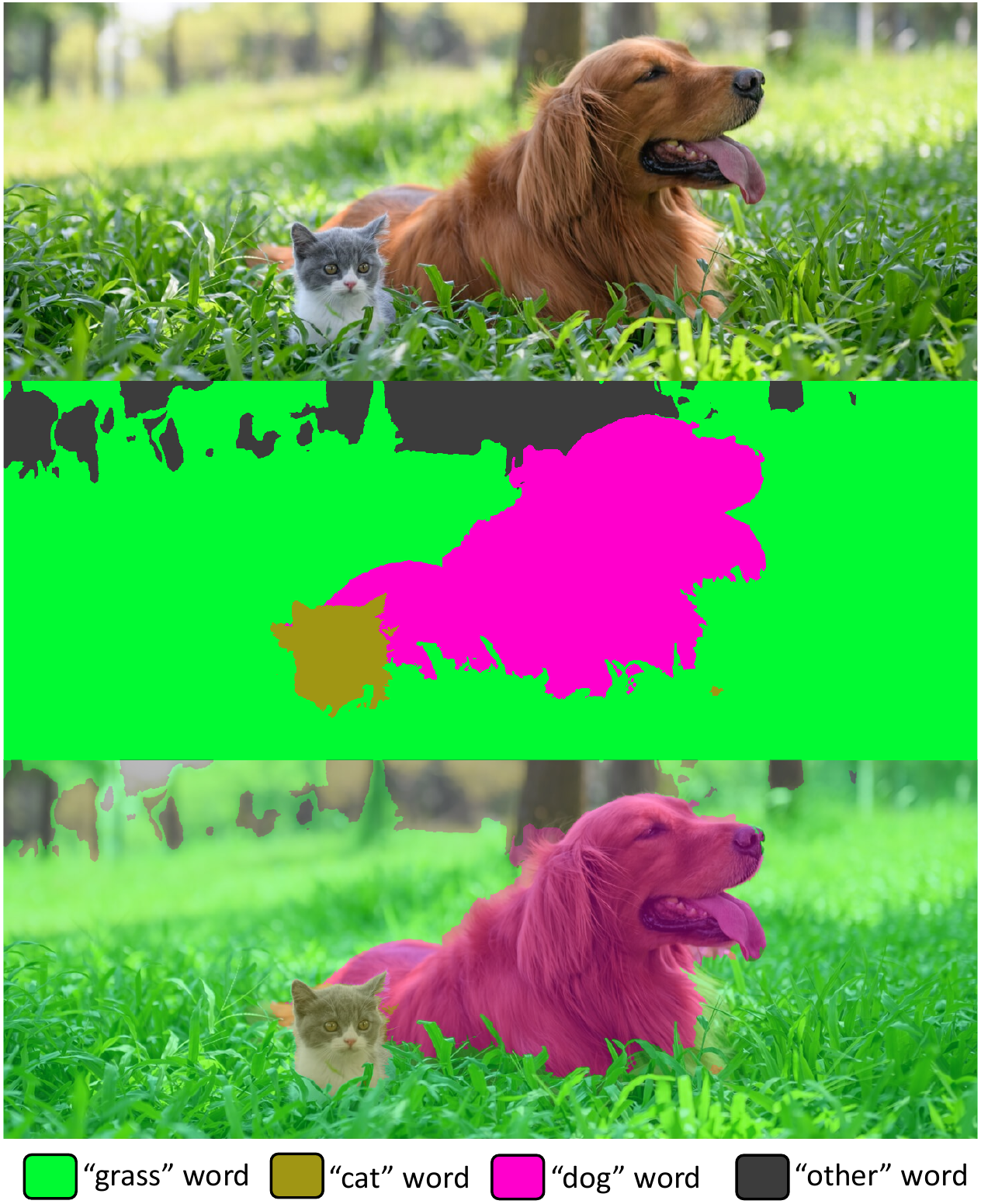}
   \caption{\textbf{Visualization of image-free segmentation results via IFSeg on a web image.} 
   Here, we present a web image (\textbf{Top}) and its segmentation results (\textbf{Middle} and \textbf{Bottom}) of our image-free segmentation approach. 
   Note that our model is not trained with any task-specific images and annotations, but only the text words (\emph{e.g.}, ``grass'', ``cat'', ``dog'' and ``other'') as semantic categories.}
   \label{fig1:first-page}
   \vspace{-0.2in}
\end{figure}
Understanding a new concept with less cost (\emph{e.g.}, collecting data, annotations, or training) is a challenging yet essential problem in machine learning~\cite{vinyals2016matching}.
The most common practice is fine-tuning a foundation model, pre-trained on a large amount of data~\cite{he2016deep,devlin2018bert, chen2020big, brown2020_gpt3}, for downstream tasks.
In particular, such large-scale models have shown successful adaptation to downstream tasks with only little supervision across vision~\cite{chen2020big} and language~\cite{brown2020_gpt3} domains.
Recently, pre-training approaches in the vision-language (VL) domain have also achieved remarkable results in transferring to novel tasks (\emph{e.g.}, few-shot or zero-shot transfer~\cite{snell2017prototypical}) with various elaborate designs, including modality interaction between the dual encoders~\cite{radford2021learning, jia2021scaling}, the multi-modal encoder~\cite{kim2021vilt, wang2022image}, and the encoder-decoder~\cite{alayrac2022flamingo, wang2022simvlm,cho2021unifying, wang2022ofa, tsimpoukelli2021multimodal, yang2022empirical}.

Semantic segmentation is one of the crucial tasks in computer vision that requires understanding dense representations for pixel-wise classifications.
Inspired by the success of the contrastive VL pre-training, CLIP~\cite{radford2021learning}, several recent attempts~\cite{zhou2022extract, li2022languagedriven, ghiasi2022scaling, xu2022simple, liu2022open} have explored CLIP-based segmentation approaches for better transferability (\emph{e.g.}, zero-shot~\cite{xian2019semantic,bucher2019zero} and open-vocabulary segmentation~\cite{zhao2017open}).
However, the existing zero-shot or open-vocabulary segmentation approaches still suffer from a burden of training on additional image data, segmentation annotations~\cite{li2022languagedriven, ghiasi2022scaling, xu2022simple, zhou2022extract}, or natural language supervision~\cite{liu2022open, xu2022groupvit}, to adapt pre-trained VL models to downstream segmentation tasks. In the wild, however, such training data is not
readily available; \emph{e.g.}, there would be no task-specific training images or labels for novel web images like \cref{fig1:first-page}.
This limitation inspires us to investigate {how to fully utilize the VL models for semantic segmentation in a lightweight manner, even without any image data or human-annotated supervision}.

Meanwhile, the recent encoder-decoder VL models~\cite{alayrac2022flamingo, wang2022simvlm,cho2021unifying, wang2022ofa, tsimpoukelli2021multimodal, yang2022empirical} also have gained popularity with their unique characteristics of image-to-text generation via the VL decoder network.
Motivated by this, we explore the potential usability of the VL decoder to segment pixels in the text generation manner as an alternative to traditional vision segmentation decoders,
\emph{e.g.}, Semantic FPN~\cite{kirillov2019panoptic} and UperNet~\cite{xiao2018unified}.
Interestingly, we found that \emph{a solely given set of semantic categories enables the encoder-decoder VL models to perform semantic segmentation without any training images or annotations};
\cref{fig1:first-page} shows the quality of semantic segmentation results on the image-free segmentation task with a wild uncurated image downloaded from the web.

\begin{figure*}[h!]
\centering
\includegraphics[width=0.95\textwidth]{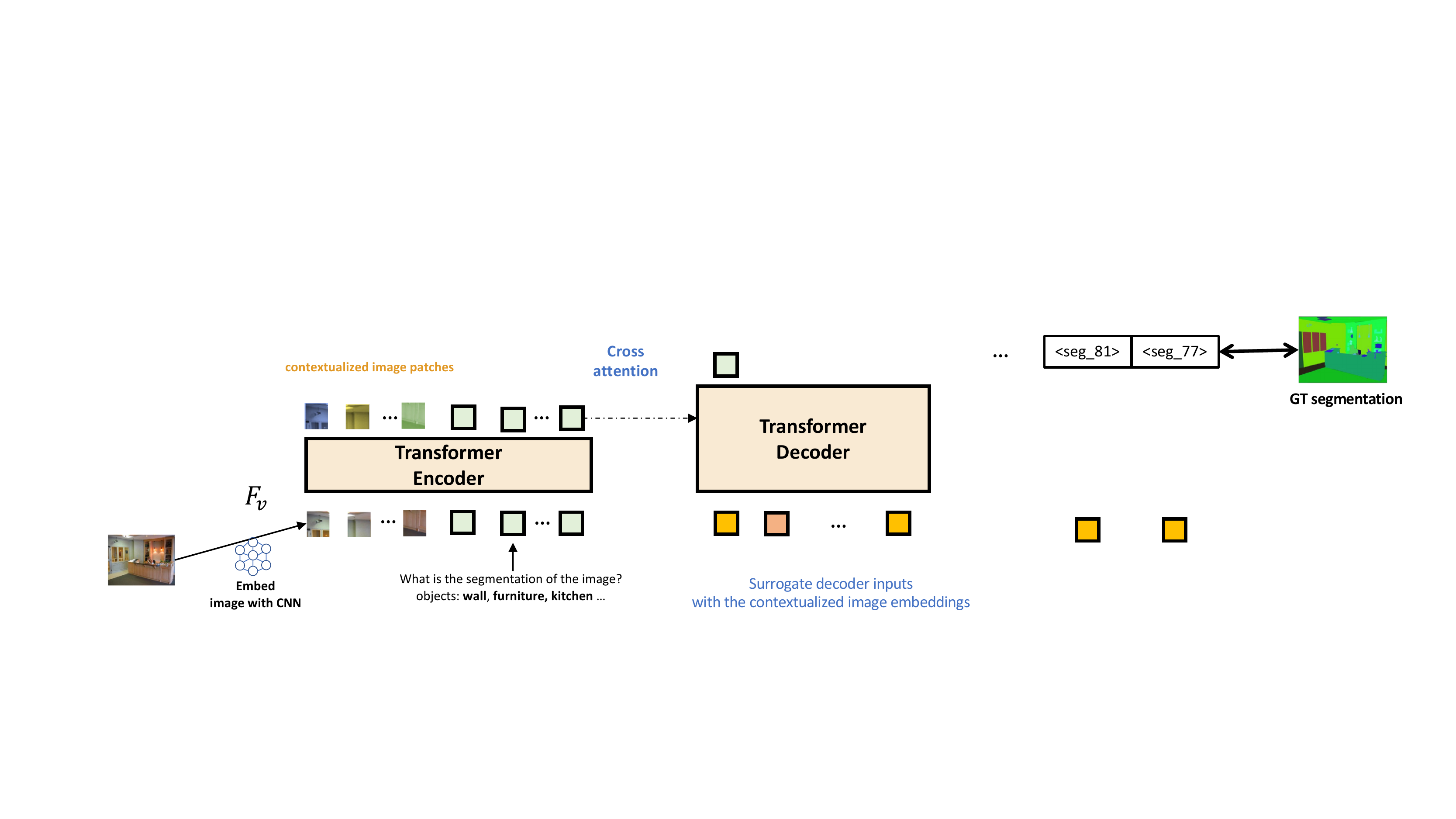}
\vspace{0.05in}
\caption{\textbf{Illustration of the semantic segmentation in VL encoder-decoder.} 
Our method incorporates a transformer encoder-decoder ($f_\mathtt{enc}, f_\mathtt{dec}$) along with an external image backbone ($f_\mathtt{img}$) for tokenizing a given image. Given a pair of an image and a prompt sentence, the transformer generates contextualized embeddings through its self-attention layers. The decoder then sequentially predicts the probability distribution over the semantic categories in a region (\emph{e.g.,} $\mathbf{p}^{(i)}$), by transforming an input composed of the special begin-of-sequence (BOS) embedding and the contextualized embeddings at the preceding region indices (\emph{e.g.,} $[\mathbf{e}_\mathtt{BOS};{f}^{(0)}(\mathbf{e}_\mathtt{x});...;{f}^{(i-1)}(\mathbf{e}_\mathtt{x})]$) through its self-attention and cross-attention layers. Finally, bilinear interpolation is applied to obtain the final prediction in a desired spatial size.
} \label{fig2:segmentation_overview}
\vspace{-0.1in}
\end{figure*}

\vspace{0.02in}
\noindent
{\bf Contribution.} 
In this paper, we introduce a novel \textbf{I}mage-\textbf{F}ree \textbf{Seg}mentation task that aims to segment target semantic categories when only a set of the target semantic categories is given without any task-specific images and annotations.
Our core idea to tackle this challenge is that a word set of semantic categories can serve as an artificial image for the VL models on their cross-modal embedding space.
To this end, we propose a simple yet effective VL-driven self-supervised task, coined \emph{IFSeg}, that generates artificial image-segmentation pairs using word tokens and updates the VL models to segment them.
Specifically, we construct this artificial training data by creating a 2D map of random semantic categories (\emph{i.e.}, artificial image tokens) and another map of their corresponding word tokens.
We provide overall illustrations and the proposed method for semantic segmentation via the VL models in \cref{fig2:segmentation_overview,fig3:ifseg_overview}, respectively.

To demonstrate the effectiveness of our method for image-free semantic segmentation, we incorporate our method with the publicly available encoder-decoder VL model~\cite{wang2022ofa}.\footnote{
Our framework can be 
incorporated with any encoder-decoder VL models
and is expected to be improved by using even larger or better VL models, \emph{cf.},
pretraining OFA was performed on 22M image-text pairs, while the popular CLIP~\cite{radford2021learning} was pre-trained on 400M image-text pairs.}
In particular, the proposed method, albeit with weaker supervision (\emph{i.e.}, only segmentation categories), can even outperform the baselines that use much stronger supervision, such as task-specific images 
and segmentation masks.
For example, our method outperforms
MaskCLIP+~\cite{zhou2022extract} without 118k training images 
on a zero-shot segmentation scenario in the COCO Stuff benchmark
by achieving
+6.9 higher mIoU.
In addition, we conduct conventional scenarios having images and annotations available for further analysis, including supervised and semi-supervised approaches.
As a result, we demonstrate our method still outperforms the recent VL-driven supervised segmentation baselines. For example, our method has achieved an improved +2.0 mIoU compared to DenseCLIP~\cite{rao2022denseclip} on the ADE20K benchmark.

Overall, our work newly introduces image-free semantic segmentation, a challenging yet potentially crucial task for the computer vision domain, and also highlights the broad applicability of the recent tending VL models.
We hope our work could inspire researchers to
rethink a new research direction for segmentation tasks in a dataset-free manner.

\section{Method}
In this section, we present a method for performing semantic segmentation tasks using vision-language (VL) encoder-decoder models and our image-free approach in a self-supervised manner. 
Inspired by the success of zero-shot transfer (\emph{e.g.}, zero-shot image classification~\cite{radford2021learning}) in the recent VL models, we aim to perform semantic segmentation only given a set of target semantic categories but without any task-specific images and annotations during training.
However, several prior works~\cite{zhou2022extract, ghiasi2022scaling} observed that it is challenging to directly segment semantic categories via VL models, \emph{e.g.}, CLIP~\cite{radford2021learning}, without any modifications and additional training.
Nonetheless, we address this challenging task using the pre-trained VL models with an encoder-decoder architecture.
In \cref{sec2:preliminaries}, we introduce the VL encoder-decoder architecture and describe how it operates in our method.
In \cref{sec2:segmentation}, we describe how the semantic segmentation task can be handled in the encoder-decoder VL model. In \cref{sec2:ifseg}, we present our image-free semantic segmentation method.

\subsection{VL Encoder-Decoder Architecture} \label{sec2:preliminaries}
Here, we introduce the VL model architecture in our framework and describe its operation step-by-step.

\vspace{0.02in}
\noindent \textbf{Data format. } Our method operates based on sequence data. For instance, let $\mathbf{x}$ be a sequence data of length $L_{\mathtt{x}}$ and let $\mathbf{e}_\mathtt{x}$ be its embedding in a $D$-dimensional vector space:
\begin{gather}
\mathbf{x} = \{{x}^{(0)},...,{x}^{(L_{\mathtt{x}}-1)}\}, \label{eq:x-sequence} \\
\mathbf{e}_\mathtt{x} = [{\mathbf{e}_\mathtt{x}}^{(0)};...;{\mathbf{e}_\mathtt{x}}^{(L_{\mathtt{x}}-1)}] \in \mathbb{R}^{L_{\mathtt{x}} \times D}.\label{eq:x-embedding}
\end{gather}
Specifically, we deal with the raw image-text $(\mathcal{X}_\mathtt{I}, \mathcal{X}_\mathtt{T})$ by tokenizing them into a sequence of tokens. The text $\mathcal{X}_\mathtt{T}$ is tokenized by a dictionary $\mathcal{V} = \{v_0, ..., v_{N-1} \}$ of $N$ pre-defined words\footnote{We utilize the bytes pair encoding (BPE) \cite{sennrich2016neural} words.} and the corresponding word embedding matrix $\mathbf{E} = [\mathbf{e}_0;...; \mathbf{e}_{N-1}] \in \mathbb{R}^{N \times D}$ that are related by the lookup operation $\mathbf{e}_{i} := \mathtt{Emb}({v}_{i})$. For example, we consider the following source text tokens and their embedding,
\begin{gather}
\mathbf{x}_\mathtt{T} = \{{x}_{\mathtt{T}}^{(0)},...,{x}_{\mathtt{T}}^{(L_{\mathtt{T}}-1)}\},\label{eq:text-tokens} \\
\mathbf{e}_\mathtt{T} = [\mathbf{e}_{\mathtt{T}}^{(0)};...;\mathbf{e}_{\mathtt{T}}^{(L_{\mathtt{T}}-1)}] \in \mathbb{R}^{L_{\mathtt{T}} \times D},\label{eq:text-embedding}
\end{gather}
where ${x}_{\mathtt{T}}^{(i)} \in \mathcal{V}$ and $\mathbf{e}_{\mathtt{T}}^{(i)} := \mathtt{Emb}({x}_{\mathtt{T}}^{(i)})$. To deal with the image $\mathcal{X}_{\mathtt{I}}$, an image backbone\footnote{Typical vision models (\emph{e.g.}, convolutional neural nets) are used.} is introduced to produce a 2D feature map of shape {$H \times W \times C$}, followed by a spatial flatten operation ($H \times W \to L_{\mathtt{I}}$), resulting in the sequence
\begin{align}
    f_\mathtt{img} (\mathcal{X}_{\mathtt{I}})=\widetilde{\mathbf{e}}_{\mathtt{I}}=[\widetilde{\mathbf{e}}_{\mathtt{I}}^{(0)};...;\widetilde{\mathbf{e}}_{\mathtt{I}}^{(L_\mathtt{I}-1)}] \in \mathbb{R}^{L_\mathtt{I} \times C}.\label{eq:image-embedding}
\end{align}
Additionally, a learnable linear layer is applied to fix the output channel size, $\mathbf{e}_{\mathtt{I}} = \mathtt{Linear}(\widetilde{\mathbf{e}}_{\mathtt{I}}) \in \mathbb{R}^{L_\mathtt{I} \times D}$, which we interpret as the embedding of the conceptual image tokens:
\begin{align}
\mathbf{x}_{\mathtt{I}}=\{x_{\mathtt{I}}^{(0)};...;x_{\mathtt{I}}^{(L_\mathtt{I}-1)}\}. \label{eq:image-tokens}
\end{align}
Concatenating them together, we assign the token sequence $\mathbf{x}:=\{\mathbf{x}_{\mathtt{I}}, \mathbf{x}_{\mathtt{T}} \}$ in \cref{eq:x-sequence} and the embedding representation $\mathbf{e}_\mathtt{x}=[\mathbf{e}_{\mathtt{I}}; \mathbf{e}_{\mathtt{T}}] \in \mathbb{R}^{L_\mathtt{x} \times D}$ in \cref{eq:x-embedding}, where $L_\mathtt{x}:=L_\mathtt{I}+L_\mathtt{T}$.

\begin{figure*}[t]
\vspace{-0.1in}
\centering
    \includegraphics[width=0.93\textwidth]{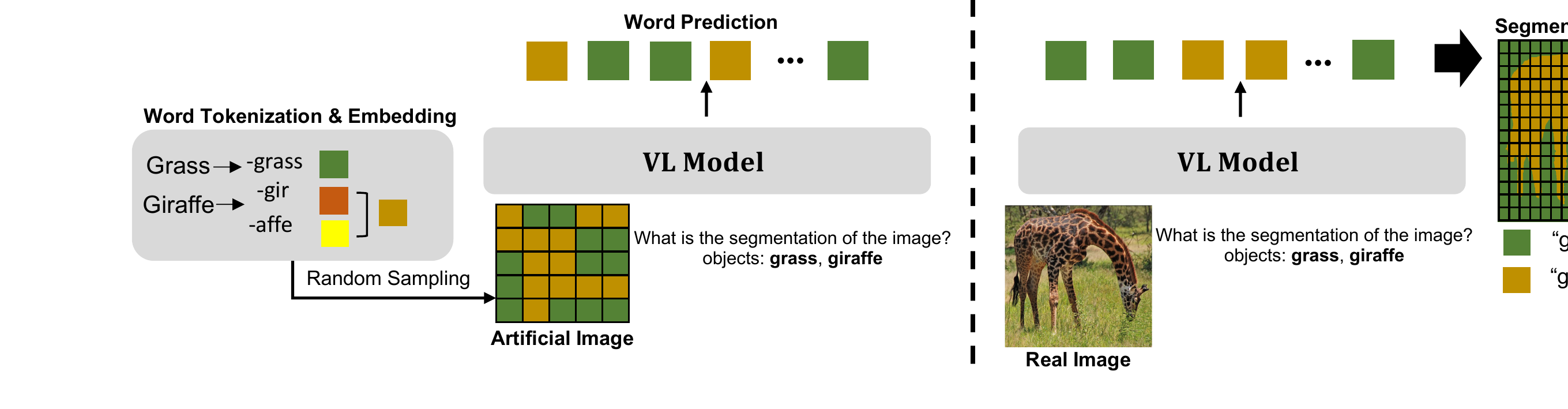}
\caption{\textbf{Overview of the proposed Image-Free Segmentation (IFSeg) task.} 
\textbf{(a) Training:} Artificial training data is constructed by randomly sampling words from the segmentation vocabulary $\mathcal{V}_\mathtt{seg}=\{v_0, v_1\}$ (\emph{e.g.}, ``$v_0$: grass'' and ``$v_1$: giraffe''). Sub-word tokens (\emph{e.g.}, ``-gir'' and ``-affe'') are managed by averaging their embeddings. Given the artificial image token $\mathbf{x}_\mathtt{I}$ and the prompt $\mathbf{x}_\mathtt{T}$, we adapt a pre-trained VL encoder-decoder to predict the corresponding word for each region of the artificial image token in a self-supervised manner (\emph{i.e.}, $\mathbf{y}_\mathtt{gt}=\mathbf{x}_\mathtt{I}$).
\textbf{(b) Inference:} During the inference on a real image $\mathcal{X}_\mathtt{I}$, the real image token is generated using the image backbone $f_\mathtt{img}(\mathcal{X}_\mathtt{I})$. The adapted VL encoder-decoder predicts the semantic category words for individual image regions (or pixels).}
\label{fig3:ifseg_overview}
\vspace{-0.15in}
\end{figure*}
\vspace{0.02in}
\noindent \textbf{VL model architecture.}
VL models predict
a target $\mathbf{y}=\{y^{(0)},...,y^{(L_\mathtt{y}-1)}\}$ based on a learned distribution $P(\mathbf{y} | \mathbf{x})$ given the multi-modal source $\mathbf{x}$. To be specific, we employ an encoder-decoder model\cite{sutskever2014sequence}, where an encoder produces a contextualized encoding of $\mathbf{x}$, and a decoder predicts the target distribution based on the encoding.
Specifically, the transformer architecture
\cite{vaswani2017attention, dosovitskiy2021vit} is adopted for implementing the modules, $f_\mathtt{enc}$ and $f_\mathtt{dec}$. The transformer encoder $f_\mathtt{enc}$ produces the contextualized embedding of $\mathbf{x}$ by transforming the embedding $\mathbf{e}_\mathtt{x}$ with the self-attention mechanism \cite{vaswani2017attention},
\begin{align}
    f_\mathtt{enc} (\mathbf{e}_\mathtt{x})
    = [ f^{(0)}_\mathtt{enc} ( \mathbf{e}_\mathtt{x} );...; f_\mathtt{enc}^{(L_{\mathtt{x}}-1)} ( \mathbf{e}_\mathtt{x} ) ] \in \mathbb{R}^{L_{\mathbf{x}} \times D}.\label{eq:transformer_encoder}
\end{align}
Then, the transformer decoder $f_\mathtt{dec}$ sequentially produces the output, by transforming a decoder input $\mathbf{d}_{i}=[\mathbf{d}^{(0)};...;\mathbf{d}^{(i)}] \in \mathbb{R}^{(i+1) \times D}$ with the self-attention and the cross-attention \cite{vaswani2017attention} mechanism with respect to $f_\mathtt{enc} (\mathbf{e}_\mathtt{x})$,
\begin{align}
\mathbf{h}^{(i)} 
= f_\mathtt{dec} (\mathbf{d}_{i}; f_\mathtt{enc} (\mathbf{e}_\mathtt{x})) \in \mathbb{R}^{D}.
\label{eq:transformer_decoder}
\end{align}
The formulation of the decoder input $\mathbf{d}_{i}$ would vary depending on the tasks. For example, the formulation during the pre-training is often the earlier targets, $\mathbf{d}^{(i)} := \mathtt{Emb} (y^{(i-1)})$ for $i>0$, and a special begin-of-sequence embedding $\mathbf{d}^{(0)} := \mathbf{e}_{\mathtt{BOS}}$. However, we will revisit and alter this formulation in \cref{sec2:segmentation} for the semantic segmentation task.

Finally, a linear transform
by the embedding matrix $\mathbf{E}$ produces a logit over the dictionary $\mathcal{V}$,
\begin{align}
{P}({y}^{(i)}|\mathbf{x}) \propto
\mathbf{E} \cdot \mathbf{h}^{(i)} \in \mathbb{R}^{N}. \label{eq:output-probability}
\end{align}
During the VL pre-training (\emph{e.g.}, image captioning),
all modules are trained end-to-end by maximizing the likelihood in \cref{eq:output-probability}.
We assume that \textit{the VL pre-training would align the image tokens with the word tokens in the contextualized embedding space} in \cref{eq:transformer_encoder}, which is the key idea in our framework introduced in \cref{sec2:ifseg}.

\subsection{Semantic Segmentation via Encoder-Decoder}\label{sec2:segmentation}
In this section, we formulate the semantic segmentation task in the VL encoder-decoder model and discuss the technical considerations. An overall pipeline is depicted in \cref{fig2:segmentation_overview}.

\vspace{0.02in}
\noindent \textbf{Task formulation.}\label{sec2:pipeline}
Given $M$ semantic categories of interest, we formulate a semantic segmentation task as decoding a category word for each dense region of the image. However, this design could be cumbersome in practice, since a certain semantic category word may be tokenized to multiple sub-words in the dictionary $\mathcal{V}$ (\emph{e.g.}, ``giraffe'' is tokenized to 2 sub-words: ``\_gir'' and ``affe'' in \cref{fig3:ifseg_overview}). As a remedy, we treat such a category as a temporary additional word and append the average embedding of the sub-word tokens to the embedding matrix $\mathbf{E}$. In this way, each semantic category is always treated as one distinct word, $\mathcal{V}_{\mathtt{seg}}=\{v'_{0},...,v'_{M-1}\}$.

To perform the task, we aim to produce spatially conditioned\footnote{We also replace the decoder's position embedding with the encoder's image position embedding for better visual understanding.} decoder outputs on the image tokens $x^{(i)}_{\mathtt{I}}$ (\emph{i.e.,} \cref{eq:image-tokens}). Specifically,
we enforce 
an alternative formulation of decoder input $\mathbf{d}_i$ in \cref{eq:transformer_decoder} such that the encoder output of the preceding index is used, \emph{i.e.}, $\mathbf{d}^{(i)} =  f^{(i-1)}_\mathtt{enc} ( \mathbf{e}_\mathtt{x} )$ for $i > 0$, where $\mathbf{d}^{(0)}=\mathbf{e}_{\mathtt{BOS}}$ without modification.
Then, we get $L_{\mathtt{I}}$ number of decoder outputs as
\begin{align}
\mathbf{h} =[\mathbf{h}^{(0)};...;\mathbf{h}^{(L_{\mathtt{I}}-1)}] \in \mathbb{R}^{L_{\mathtt{I}} \times D}. \label{eq:segmentation-latent-initial}
\end{align}
Next, we calculate the logit with \cref{eq:output-probability} and apply softmax after masking out the words that are not in $\mathcal{V}_{\mathtt{seg}}$ to get the normalized probability over the $M$ categories,
\begin{align}
\mathbf{p} =[\mathbf{p}^{(0)};...;\mathbf{p}^{(L_{\mathtt{I}}-1)}] \in \mathbb{R}^{L_{\mathtt{I}} \times M}. \label{eq:segmentation-probability-initial}
\end{align}
Then, we recover the spatial dimension of the image backbone $f_{\mathtt{img}}$ (\emph{i.e.}, $L_{\mathtt{I}} \to H \times W$) and up-sample it with bilinear interpolation to match a desired size $\widetilde{P} \times \widetilde{W}$ (\emph{e.g.}, an irregular shape of the image $\mathcal{X}_\mathtt{I}$). As a result, we obtain the output
\begin{align}
    \widetilde{\mathbf{p}} = [\widetilde{\mathbf{p}}^{(0)};...;\widetilde{\mathbf{p}}^{(\widetilde{H}\cdot \widetilde{W}-1)}] \in \mathbb{R}^{\widetilde{H} \times \widetilde{W} \times M}\label{eq:segmentation-probability-upsample},
\end{align}
and the predictive distribution is defined as:
\begin{align}
P(y^{(i)}|\mathbf{x}) := \widetilde{\mathbf{p}}^{(i)} \in \mathbb{R}^{M}.
\end{align}
Finally, we predict the category with the highest probability,
\begin{align}
\hat{y}^{(i)} = \argmax_{y \in \mathcal{V_{\mathtt{seg}}}}{P({y}^{(i)}=y | \mathbf{x})}. \label{eq:segmentation-prediction}
\end{align}
For fine-tuning given a segmentation label $y^{(i)}_{\mathtt{gt}}$ (represented by the semantic category words in $\mathcal{V_{\mathtt{seg}}}$), we consider the negative log-likelihood as the objective to minimize:
\begin{align}
\mathcal{L}_{\mathtt{seg}}(\mathbf{x}, \mathbf{y}_\mathtt{gt}) = \sum_{i} -\ln {P}(y^{(i)}={y}_{\mathtt{gt}}^{(i)}|\mathbf{x}).\label{eq:finetune-objective}
\end{align}

\vspace{0.02in}
\noindent \textbf{Prompt design.}
The text tokens $\mathbf{x}_\mathtt{T}$ in \cref{eq:text-tokens} can be provided as the prompt for instructing the details of the semantic segmentation task, namely the task description and the list of target classes. Specifically, we follow the ``\textit{task description} $+$ \textit{category enumeration}'' protocol in the VQA task~\cite{wang2022ofa} where the target classes are enumerated after the task description, \emph{e.g.}, ``what is the segmentation map of the image? object: giraffe, grass,'' in \cref{fig3:ifseg_overview}.
In this design, we expect the VL model to capture the cross-modal relationships between image tokens $\mathbf{x}_\mathtt{I}$ and the semantic categories.

\subsection{Image-free Semantic Segmentation}\label{sec2:ifseg}
In this section, we introduce a VL-driven self-supervised task, coined \emph{IFSeg} (\textbf{I}mage-\textbf{F}ree \textbf{Seg}mentation), to tackle the image-free semantic segmentation via the encoder-decoder VL model. 
Our main idea is that during the VL pre-training (in \cref{sec2:preliminaries}), the real image tokens and their corresponding semantic category word tokens can be considered interchangeable because they are both likely to be located in close proximity within the shared contextualized embedding space.
To this end, we generate artificial image tokens using given word tokens and update the VL model to segment the corresponding word tokens in a self-supervised manner. In other words, we generate artificial training data for an image-free semantic segmentation task.
We provide a brief overview of the proposed image-free approach in \cref{fig3:ifseg_overview}.

\vspace{0.02in}
\noindent
\textbf{Constructing artificial image tokens.}
We construct artificial training data (\emph{i.e.,} image-segmentation token pairs) from a set of $M$ unique category words $\mathcal{V}_{\mathtt{seg}}:=\{v'_{0},...,v'_{M-1}\}$.
Specifically, we randomly sample with replacement $U \times V$ number of words to construct a grid map $\widetilde{\mathbf{v}}_{\mathtt{IFSeg}}$ as follows:
\begin{gather}
    \widetilde{\mathbf{v}}_{\mathtt{IFSeg}} = \{\widetilde{v}_\mathtt{IFSeg}^{(0)}, ..., \widetilde{v}_\mathtt{IFSeg}^{(U\cdot V-1)}\}.\label{eq:artificial_image_uv}
\end{gather}
The initial grid sizes $U$, $V$ are randomly drawn from a range $\{1,2, ..., S\}$ with a hyper-parameter $S$.
Then, we up-scale the grid to have the spatial resolution of the image backbone (\emph{i.e.}, $H\times W$) via the nearest neighbor interpolation,
\begin{gather}
    {\mathbf{v}}_{\mathtt{IFSeg}} = \{{v}_\mathtt{IFSeg}^{(0)}, ..., {v}_\mathtt{IFSeg}^{(H\cdot W-1)}\}.\label{eq:artificial_image_hw}
\end{gather}
In our experiments, we use $H=W=32$ 
by following the configuration of the VL pre-training, 
and we also set $S=32$ as the size of the initial map, so it
can vary in the largest range
(see \cref{supple:abl:hyperparameter} for analysis on the effect of the initial grid range $S$).
The goal of using various random maps to up-sample our data is to bridge the gap between real images and our synthetic data by introducing a shape regularization effect. This effect allows objects to be depicted as a cluster of various sizes rather than being randomly scattered.
Finally, we train the model with the artificial image tokens $\mathbf{v}_{\mathtt{IFSeg}}$ (replacing the real image tokens in \cref{eq:image-tokens}) and their corresponding ground truths using the \textit{maximum likelihood} in \cref{eq:finetune-objective}.
We note that the image backbone, $f_\mathtt{img}$ (in \cref{eq:image-embedding}) is frozen during our self-supervised training.

\vspace{0.02in}
\noindent
\textbf{Post-processing for image-free segmentation.}
One challenge of the image-free segmentation task is the discrepancy in input modality between training and evaluation, which arises due to the absence of real training images.
For example, it is challenging to learn image-specific priors such as object shapes and label coherence in regions with similar textures.
To resolve this issue, we found that averaging the output probability based on the image feature (\emph{i.e.,} outputs of image backbone $f_\mathtt{img}$) significantly enhances the segmentation quality.
Specifically, we search $K$-nearest neighbors of the image features in \cref{eq:image-embedding} using the cosine similarity,
$\widetilde{\mathbf{e}}_{\mathtt{I}}^{(i)}
\cdot
\widetilde{\mathbf{e}}_{\mathtt{I}}^{(j)}
/
\|\widetilde{\mathbf{e}}_{\mathtt{I}}^{(i)}\|
\cdot
\|\widetilde{\mathbf{e}}_{\mathtt{I}}^{(j)}\|$.
Then, given a set of neighborhood indices $\mathcal{N}^{(i)}$, we iterate averaging the probability in \cref{eq:output-probability} with the neighborhood as follows,
\begin{align}
\mathbf{p}^{(i)}:=\sum_{j\in \mathcal{N}^{(i)}} \mathbf{p}^{(j)} / \:|\mathcal{N}^{(i)}|.
\end{align}
We empirically found that the effect of the post-processing diminishes when the real training images and annotations are available.
In our experiments, we apply this only for image-free approaches and use $K=3$ and 25 iterations unless stated otherwise (see \cref{supple:abl:hyperparameter} for ablation studies on varying $K$ and the iteration count).
\section{Related Works}
\noindent {\bf Vision-language pre-training.}
The recent vision-language models pre-trained on large-scale image-text data have shown successful results in zero-shot and few-shot adaptation to novel tasks across domains, \emph{e.g.}, image classification~\cite{deng2009imagenet}, captioning~\cite{lin2014microsoft} and visual question answering~\cite{antol2015vqa}.
To improve the quality of cross-modal representations, there have been extensive exploration in design of modality interaction, including the dual encoders~\cite{radford2021learning, jia2021scaling}, the multi-modal encoder~\cite{kim2021vilt, wang2022image}, and the encoder-decoder~\cite{alayrac2022flamingo, wang2022simvlm,cho2021unifying, wang2022ofa, tsimpoukelli2021multimodal, yang2022empirical}.
For example, CLIP~\cite{radford2021learning} introduced contrastive pre-training on the dual encoder (\emph{i.e.}, image and text encoder) and has shown impressive zero-shot image classification performances via a simple prompt engineering technique without training.
On the other hand, the encoder-decoder VL approaches~\cite{alayrac2022flamingo, wang2022simvlm,cho2021unifying, wang2022ofa, tsimpoukelli2021multimodal, yang2022empirical} also have gained much attention in image-to-text generation tasks such as image captioning and visual question answering. 
In this paper, we explore the potential usability of the VL decoder for image segmentation from the perspective of image-to-text generation.

\vspace{0.02in}
\noindent
{\bf Transferable image segmentation.}
Image segmentation is a core computer vision task, but it is still challenging to segment novel visual categories.
To this end, several attempts have been introduced, including unsupervised~\cite{yin2022transfgu, hamilton2022unsupervised, ji2019invariant, cho2021picie, liu2022open, zhou2022extract} and zero-shot segmentation~\cite{gu2020context, bucher2019zero, xian2019semantic, cheng2021sign, zhou2022extract, li2022languagedriven, ghiasi2022scaling, xu2022simple, pastore2021closer}.
First, unsupervised segmentation approaches~\cite{yin2022transfgu, hamilton2022unsupervised, ji2019invariant, cho2021picie, zhou2022extract} have been focused on clustering dense representations of an image, and then matching corresponding segmentation categories via the Hungarian-matching algorithm~\cite{doersch2015unsupervised}. 
On the other hand, the recent VL-driven approaches~\cite{liu2022open, zhou2022extract} replace the matching process via the text encoder of CLIP using segmentation vocabulary for better efficiency and transferability.
Meanwhile, early approaches in zero-shot segmentation~\cite{gu2020context, bucher2019zero, xian2019semantic, cheng2021sign, pastore2021closer} have utilized segmentation vocabulary via learned word embeddings like word2vec~\cite{mikolov2013distributed} and fast-text~\cite{joulin2016fasttext}.
Similar to the VL-driven unsupervised segmentation, the VL-driven zero-shot approaches~\cite{zhou2022extract, li2022languagedriven, ghiasi2022scaling, xu2022simple} also have been established on CLIP instead of word embeddings.
The zero-shot segmentation approaches often require class-agnostic segmentation masks~\cite{ghiasi2022scaling, xu2022simple} or class-specific segmentation annotations~\cite{gu2020context, bucher2019zero, xian2019semantic, cheng2021sign, li2022languagedriven, zhou2022extract, pastore2021closer}.
In this respect, we explore an image-free semantic segmentation task for more realistic scenarios with only given segmentation vocabulary, which can be easily collected than images or other annotations.
\begin{figure*}
\vspace{-0.1in}
 \centering
  \includegraphics[width=0.93\textwidth]{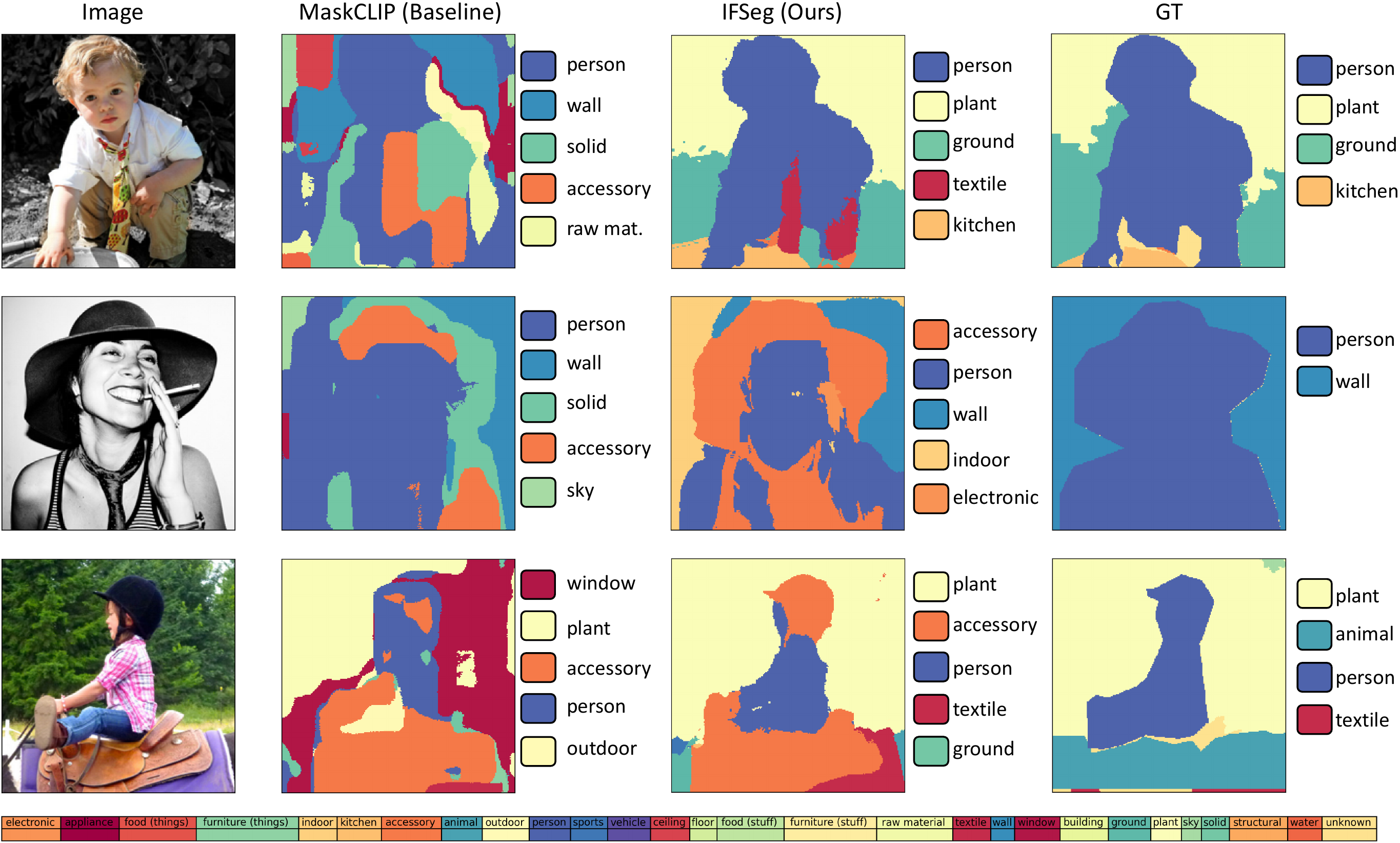}
\caption{\textbf{Visualization of segmentation results via IFSeg.}
We visualize the segmentation results of IFSeg (ours) and MaskCLIP (baseline). 
We also present predicted semantic categories next to each segmentation results. 
Unlike the MaskCLIP (baseline) only roughly segments segmentation vocabularies onto an image, our method does visual categories with accurate segmentation.
We note that both models are not trained using any images from the pre-trained VL models, CLIP and OFA, respectively.
Best viewed in color.} \label{fig4:visualize_ifseg}
\vspace{-0.1in}
\end{figure*}
\begin{table}[t]
\centering
\small
\scalebox{1.0}
{
\begin{tabular}{llcc}
    \toprule
    Method & Backbone & Image Dataset & mIoU \\
    \midrule
    MaskCLIP+ \cite{zhou2022extract} & ResNet-101 & COCO (118k) & 48.7 \\
    \midrule
    CLIP \cite{radford2021learning,zhou2022extract} & ResNet-101 & \xmark & 12.3 \\
    OFA \cite{wang2022ofa}  & ResNet-101  & \xmark & {6.8} \\ 
    MaskCLIP\cite{zhou2022extract}  & ResNet-101 & \xmark & 24.8 \\
    IFSeg (ours) & ResNet-101  & \xmark & \textbf{55.6} \\
    \bottomrule
\end{tabular}
}
\caption{\textbf{Comparison with zero-shot and image-free baselines.} We report the mIoU metric of the baselines and our model predicting the 15 unseen semantic categories of the COCO Stuff benchmark.
 “Image Dataset” denotes required images for training.
Our post-processing has been applied to all results for a fair comparison.}
\label{tbl1:coco_unseen}
\vspace{-0.1in}
\end{table}

\section{Experiments}
\label{sec:experiments}
In this section, we demonstrate the effectiveness of the proposed image-free approach, IFSeg. Specifically, we incorporate our method with the recent VL encoder-decoder model, OFA~\cite{wang2022ofa}, which is publicly available,\footnote{\url{https://github.com/OFA-Sys/OFA}.} and evaluate its segmentation abilities on COCO Stuff~\cite{caesar2018coco} and ADE20K~\cite{zhou2017scene} semantic segmentation benchmarks.
Specifically, we compare our method with existing VL-driven segmentation baselines that target various scenarios: (a) zero-shot segmentation scenario~\cite{gu2020context, bucher2019zero, xian2019semantic, cheng2021sign, zhou2022extract, ghiasi2022scaling, xu2022simple}, (b) cross-dataset segmentation scenario~\cite{li2022languagedriven, ghiasi2022scaling, xu2022simple} and (c) unsupervised image segmentation~\cite{yin2022transfgu, hamilton2022unsupervised, ji2019invariant, cho2021picie, zhou2022extract}.
We consider CLIP~\cite{radford2021learning}, MaskCLIP~\cite{zhou2022extract}, and OFA~\cite{wang2022ofa} as baselines to evaluate the segmentation abilities of the pre-trained VL models without fine-tuning.
More details are described in each section and Appendix.

\vspace{0.02in}
\noindent
{\bf Datasets.}
COCO Stuff~\cite{caesar2018coco} is a large-scale dataset that contains 117k training, 5k validation images, and segmentation annotations of 171 semantic categories consisting of 80 objects and 91 stuff categories.
For the zero-shot image segmentation, we split COCO Stuff dataset into 156 seen categories and 15 unseen categories.\footnote{We report the specific vocabulary of unseen semantic categories in the COCO Stuff: \emph{frisbee, skateboard, cardboard, carrot, scissors, suitcase, giraffe, cow, road, wall concrete, tree, grass, river, clouds, playing field}.}
ADE20K~\cite{zhou2017scene} is a challenging semantic segmentation dataset including 20k training, 5k validation, and segmentation annotations of 150 fine-grained semantic categories that cover indoor and outdoor scenes.
In our image-free experiments in \cref{sec:image-free}, we use only semantic categories given by the segmentation benchmarks, without any training images and annotations.

\vspace{0.02in}
\noindent
{\bf Baselines.}
We consider a variety of existing VL-driven unsupervised, zero-shot, and the image-free segmentation baselines:
(a) unsupervised baselines:
IIC~\cite{ji2019invariant}, PiCIE+H.~\cite{cho2021picie}, TransFGU~\cite{yin2022transfgu},
(b) zero-shot baselines:
LSeg+\footnote{A re-implemented LSeg~\cite{li2022languagedriven} in the OpenSeg~\cite{ghiasi2022scaling}.}~\cite{li2022languagedriven}, ZSSeg~\cite{xu2022simple}, OpenSeg~\cite{ghiasi2022scaling}, and MaskCLIP+~\cite{zhou2022extract}, where ZSSeg, OpenSeg, and MaskCLIP+ are the recent VL-driven baselines that relied on CLIP~\cite{radford2021learning} or ALIGN~\cite{jia2021scaling},
and (c) image-free baselines: OFA~\cite{wang2022ofa}, CLIP~\cite{radford2021learning}, and MaskCLIP~\cite{zhou2022extract} which directly evaluate the segmentation abilities of the pre-trained VL models, OFA and CLIP.

\begin{table*}[t]
\centering
\small
{
\begin{tabular}{lllccc}
    \toprule
    Method & Text Backbone & Image Backbone & Image Dataset & Segmentation Label & mIoU \\
    \midrule
    LSeg+ \cite{li2022languagedriven, ghiasi2022scaling} & ALIGN-BERT-Large~\cite{jia2021scaling} & ResNet-101 & COCO (118k) & \cmark & 13.0 \\
    OpenSeg \cite{ghiasi2022scaling}& ALIGN-BERT-Large~\cite{jia2021scaling} & ResNet-101 & COCO (118k) & \cmark & 15.3 \\
    ZSSeg \cite{xu2022simple} & CLIP-ViT-B~\cite{radford2021learning} & ResNet-101 & COCO (118k) & \cmark & 20.5 \\
    \cmidrule{1-6}
    CLIP$\dag$ \cite{radford2021learning,zhou2022extract} & CLIP-ResNet~\cite{radford2021learning} & ResNet-101 & \xmark & \xmark & 3.7 \\
    MaskCLIP$\dag$ \cite{zhou2022extract} & CLIP-ResNet~\cite{radford2021learning} & ResNet-101 & \xmark & \xmark & 10.3 \\
    OFA$\dag$ \cite{wang2022ofa} & OFA-Base~\cite{wang2022ofa} & ResNet-101 & \xmark & \xmark & {0.5} \\
    IFSeg (ours)$\dag$ & OFA-Base~\cite{wang2022ofa} & ResNet-101 & \xmark & \xmark & \textbf{16.8} \\
    \bottomrule
\end{tabular}
}
\caption{\textbf{Comparison with VL-driven baselines under the cross-dataset  (COCO$\rightarrow$ADE20K) scenario.}
We report the mIoU metric evaluated on the ADE20K benchmark. 
We use the 150 fine-grained semantic categories of the ADE20K for image-free training.
``Image Dataset'' and ``Segmentation Label'' denote requirements for their training. $\dag$ denotes results that our post-processing is applied.}\label{tbl2:cross2ade}
\end{table*}
\vspace{0.02in}
\noindent
{\bf Implementation details.}
In our experiments, we implement our method on the OFA (encoder-decoder VL model) framework and generally follow the training and evaluation configuration of OFA~\cite{wang2022ofa}, $\tt {mmsegmentation}$\footnote{\url{https://github.com/open-mmlab/mmsegmentation}.} \cite{mmseg2020}, and MaskCLIP~\cite{zhou2022extract} (the strongest baseline) for a fair comparison.
We fine-tune our model from the OFA-Base pre-trained weights with the ResNet-101 backbone network. We optimize with AdamW optimizer~\cite{loshchilov2018adamw} with a weight decay of 0.1, a learning rate of 0.00005, and a batch size of 16 with 2k iterations unless stated otherwise. We generate $32\times 32$ grid-size of artificial image tokens with $S=32$ and use $K=3$ with 25 iterations for the post-processing for image-free baselines. 
We report a single-scale mean Intersection over Union (mIoU) score evaluated at the original irregular image sizes as the metric.
More details of experimental setups are described in Appendix.

 \begin{table}[t]
\centering
\small
{
\begin{tabular}{llcc}
    \toprule
    Method & Backbone & Image Dataset & mIoU \\
    \midrule
    IIC \cite{ji2019invariant}& ResNet-18 & COCO (118k) & 0.6 \\
    PiCIE + H. \cite{cho2021picie}& ResNet-18 & COCO (118k) & 4.6  \\
    TransFGU \cite{yin2022transfgu}& ViT-S/8 & COCO (118k) & 11.9  \\
    MaskCLIP+ \cite{zhou2022extract} & ResNet-101 & COCO (118k) & {18.0} \\
    \cmidrule{1-4}
    CLIP$\dag$ \cite{radford2021learning,zhou2022extract}& ResNet-101 & \xmark & 4.6 \\
    MaskCLIP$\dag$ \cite{zhou2022extract}& ResNet-101 & \xmark & 12.7 \\
    OFA$\dag$  \cite{wang2022ofa} & ResNet-101 & \xmark & {1.5} \\ 
    IFSeg (ours)$\dag$ & ResNet-101 & \xmark & \textbf{16.9} \\ 
    \bottomrule
\end{tabular}
}
\caption{\textbf{Comparison with unsupervised semantic segmentation (COCO$\rightarrow$COCO) baselines.}
We report the mIoU metric evaluated on the 171 semantic categories of the COCO Stuff benchmark. $\dag$ denotes results that our post-processing is applied.}\label{tbl3:unsup_coco}
\end{table}

\subsection{Image-free Adaptation for Segmentation}\label{sec:image-free}
\noindent {\bf Zero-shot image segmentation.}
We first evaluate the effectiveness of the proposed image-free approach, IFSeg, for adapting VL models toward semantic segmentation tasks.
We evaluate the mIoU scores of different models on segmenting the COCO Stuff 15 unseen semantic categories.
Specifically, we compare with the image-free baselines, CLIP\cite{radford2021learning}, OFA\cite{wang2022ofa}, and MaskCLIP\cite{zhou2022extract} in \cref{tbl1:coco_unseen}. 
In addition, we also compare with MaskCLIP+\cite{zhou2022extract} under the same evaluation setup as a baseline, 
which is trained on 118k COCO images using the pseudo-labels generated by MaskCLIP\cite{zhou2022extract}.
First of all, \cref{tbl1:coco_unseen} shows that our method can achieve significant improvement in mIoU metric compared to all the image-free baselines, \emph{e.g.}, +30.8 points higher than MaskCLIP. 
Somewhat surprisingly, our method outperforms MaskCLIP+\cite{zhou2022extract}, which is a stronger baseline trained on additional 118k images, despite our scarce training data regime that does not use any images and annotations except segmentation vocabulary.

\vspace{0.02in}
\noindent
{\bf Cross-dataset transfer.}
Again, we compare with VL-driven segmentation baselines in \cref{tbl2:cross2ade} under a cross-dataset scenario,
where the model is trained on the COCO Stuff and evaluated on the ADE20K benchmark.
To this end, we train our model using segmentation vocabulary of the COCO Stuff, and then evaluated on the ADE20K vocabulary. 

Similar to 
the above zero-shot scenario,
\cref{tbl2:cross2ade} shows 
that our method can achieve significant and comparable performance with the image-free baselines and the baselines with stronger supervision despite our image-free training regime. 
For example, ours achieved 1.5 points higher mIoU than OpenSeg\cite{ghiasi2022scaling} trained on the 118k training images and class-agnostic segmentation mask annotations.
Although the reported value of ours is lower than ZSSeg\cite{xu2022simple}, we note that there exists a huge gap between training scale; ZSSeg is trained on the COCO Stuff dataset with its natural language annotations (\emph{i.e.} captions), in a total $960\times$ larger training configuration ($15\times$ larger iterations with $64\times$ larger batch size).
Nevertheless, our method still consistently and significantly outperforms all the image-free baselines by a large margin; for example, ours achieves 5.5 higher points than MaskCLIP in terms of the mIoU metric. 

\vspace{0.02in}
\noindent
{\bf Unsupervised image segmentation.}
On the other hand, we also compare our method with unsupervised segmentation baselines in \cref{tbl3:unsup_coco}, which is another promising approach for learning transferable segmentation models.
Specifically, unsupervised baselines are trained on the COCO Stuff dataset and evaluated 171 semantic categories.

As shown in \cref{tbl3:unsup_coco}, our method consistently outperforms all the existing image-free segmentation baselines. For example, our method significantly outperforms MaskCLIP by achieving 16.9 mIoU, while MaskCLIP achieves 12.7. 
Also, ours shows comparable results to MaskCLIP+, which requires additional training with 118k images for transferring the knowledge of MaskCLIP via pseudo-labeling.

\vspace{0.02in}
\noindent
{\bf Qualitative Results.}
We present visualizations of segmentation results obtained by MaskCLIP and Ours in \cref{fig4:visualize_ifseg}, and it shows that our method even segments more fine-grained categories than the ground-truth labels; for example, the accessory category in the middle and bottom images are captured via ours, but not contained in the labels.

\begin{table}[t]
\centering
\small
\scalebox{0.94}{
{
\begin{tabular}{lc|ccc}
    \toprule
    Method & ST & Image Dataset & Segmentation Label & mIoU \\
    \midrule
    IFSeg & \xmark & \xmark & \xmark & {55.6} \\ 
    IFSeg  & \cmark & COCO (118k) & \xmark & {56.2} \\
    IFSeg  & \cmark & COCO (118k) & \cmark & \textbf{61.6} \\
    \bottomrule
\end{tabular}
}
}
\caption{\textbf{Ablation study on the effect of self-training technique (``ST'') of IFSeg}. All models are evaluated on the 15 unseen categories of the COCO Stuff zero-shot segmentation benchmark. 
We show the effects of task-specific images and segmentation labels (seen) with additional 8k training iterations.}\label{tbl4:ablation}
\end{table}

\subsection{Ablation study}\label{sec:ablation}
In this section, we perform an ablation study to understand further how the proposed method works when training images or segmentation annotations are available. 

\vspace{0.02in}
\noindent \textbf{Self-training.}
Self-training technique~\cite{bucher2019zero} has been widely used in the VL literature. It generates pseudo-labels of unseen segmentation categories for reducing the gap between seen and unseen semantic categories in a semi-supervised manner; it assumes the pixels of unseen categories could be present in the training images, while those pixels are not annotated. On this line, we also evaluate our method on the COCO Stuff benchmark when training images or the seen annotations are available.
Specifically, we fine-tune IFSeg with an additional 8k training iterations using 118k images and the seen annotations. We then evaluate the model on the 15 unseen categories of the COCO Stuff benchmark.
\cref{tbl4:ablation} shows the individual effects of training images and seen annotations in our framework.
After self-training, our method has improved significantly from 55.6 to 61.6 mIoU, which also largely surpasses the strongest baseline MaskCLIP+ of 48.7 on the COCO Stuff in \cref{tbl1:coco_unseen}.
Furthermore, we observe that ours can achieve outperforming performance compared to self-training baselines as presented in \cref{supple:selftraining}.

\vspace{0.02in}
\noindent
\textbf{Supervised semantic segmentation.}
Here, we perform supervised learning on the ADE20K benchmark varying model size of OFA~\cite{wang2022ofa} to demonstrate their effectiveness. For a fair comparison, we follow the training configuration of DenseCLIP~\cite{rao2022denseclip}, which incorporates cross-modal representations of CLIP to Semantic FPN~\cite{kirillov2019panoptic}, including input resolutions, batch size, and iterations.
We also compare with traditional image segmentation decoders like Semantic FPN and UPerNet~\cite{xiao2018unified} on pre-trained ImageNet~\cite{he2016deep}. 

As shown in \cref{tbl5:supervised_ade}, the encoder-decoder VL models can be successfully fine-tuned to segment semantic categories by surpassing the existing supervised approaches with a large margin, \emph{e.g.}, + 2.0 mIoU compared to the strongest baselines, DenseCLIP, on the ADE20K benchmark.
\begin{table}[t]
\centering
\small
\begin{tabular}{llc}
    \toprule
    Method & Backbone & mIoU  \\
    \midrule
    Semantic FPN \cite{kirillov2019panoptic} & ResNet-101 & 40.4 \\
    UPerNet \cite{xiao2018unified} & ResNet-101 & 43.8 \\
    \cmidrule{1-3}
    CLIP + Semantic FPN \cite{radford2021learning, rao2022denseclip} & ResNet-101 & 42.7 \\
    DenseCLIP + Semantic FPN \cite{rao2022denseclip} & ResNet-101 & 45.1 \\
    IFSeg (ours) & ResNet-101  & \bf{47.1} \\
    \bottomrule
\end{tabular}
\caption{\textbf{Comparison in supervised semantic segmentation.} We report the mIoU metric evaluated on the 150 semantic cateogires of the ADE20K benchmark. We follow training configurations of DenseCLIP, such as image resolutions and training iterations.}\label{tbl5:supervised_ade}
\label{tab:ade-sup}
\end{table}
\section{Conclusion}
We newly introduce a novel image-free semantic segmentation task, which has the goal of performing semantic segmentation without any task-specific images and annotations, except target semantic categories.
To tackle this, we propose a simple yet effective image-free framework via vision-language (VL) models in a self-supervised manner. 
The key idea is that words of semantic categories can act as an artificial image tokens on the cross-modal representation space of pre-trained VL models.
Specifically, we generate artificial image-segmentation pairs using word tokens to replace the real image-segmentation pairs for image-free semantic segmentation via the VL models.
Through extensive experiments, we demonstrate our models are not only effective baseline for this novel task but also show strong performances over existing methods acquiring the stronger supervision.
We believe our work would provide insights into the under-explored yet important problems for semantic segmentation via the pre-trained VL models.

\vspace{0.02in}
\noindent \textbf{Acknowledgements.} This work was supported by Institute of Information \& communications Technology Planning \& Evaluation (IITP) grant funded by the Korea government (MSIT) (No.2019-0-00075, Artificial Intelligence Graduate School Program (KAIST); No.2021-0-02068, Artificial Intelligence Innovation Hub; No.2022-0-00959, Few-shot Learning of Casual Inference in Vision and Language for Decision Making).

%%%%%%%%% REFERENCES
{\small
\bibliographystyle{ieee_fullname}
\bibliography{egbib}
}

\newpage
\onecolumn
\clearpage
\begin{center}{\bf {\LARGE Supplementary Material}}
\end{center}
\vspace{0.05in}
\appendix

\section{More experiments with stronger supervision}
In this section, we consider two additional scenarios when a stronger level of supervision is available; external semantic categories (see \cref{sec:larger-set}), external training images and annotations (see \cref{sec:external-images}).
Lastly, we present a comparison with (weakly) supervised baselines using the self-training technique (see \cref{supple:selftraining}).

\subsection{External segmentation categories}\label{sec:larger-set}
We investigate the effect of (a) hierarchical semantic categories and (b) external semantic categories from other sources.

\vspace{0.02in}
\noindent
\textbf{Hierarchical semantic categories.} 
Hierarchical semantic categories can be a stronger supervision for our artificial image creation.
Specifically, we explore the semantic hierarchy as the language by supervising the model with multiple example words (\emph{i.e.}, fine-grained categories) per single semantic category (\emph{i.e.}, coarse categories). 
To this end, we first introduce the COCO Stuff benchmark~\cite{caesar2018coco} with the 27 coarse semantic categories, which remaps the original 171 fine-grained categories in the COCO stuff benchmark to the 27 coarse categories.\footnote{The full list of hierarchy between the coarse and fine-grained categories are given in \cref{tbl:coco-coarse-mapping}.} 
Then we augment each coarse category's words with those from its fine-grained categories for generating the artificial image;
we slightly alter the artificial image creation in \cref{sec2:ifseg} to sample $h \cdot w$ coarse categories first, then perform additional sampling that actually assigns a word among the fine-grained categories associated with each coarse category. 
In our experiments, we empirically found that such hierarchical supervision significantly improves the performance of our method from 21.2 to 31.0 (+ 9.8) mIoU on the 27 coarse categories of the COCO Stuff benchmark.
Furthermore, we provide a comparison with unsupervised semantic segmentation baselines on the coarse COCO Stuff benchmark. 
\cref{tbl:coco-word-augmentation} summarizes the results; our method consistently and significantly outperforms all the existing baselines. For example, our method significantly outperforms STEGO~\cite{hamilton2022unsupervised} by achieving 31.0 mIoU in an image-free manner, while STEGO does 26.8, despite it requires task-specific images for training.

\begin{table*}[h!]
\centering
\small
\begin{tabular}{l c l c c}
    \toprule
    Model & Text Backbone & Image Backbone & Image Dataset & mIoU \\
    \midrule
    IIC \cite{ji2019invariant} & \xmark & ResNet-18 & COCO (118k) & 2.4  \\
    PiCIE + H. \cite{cho2021picie} & \xmark & ResNet-18 & COCO (118k) &  14.4  \\
    TransFGU \cite{yin2022transfgu} & \xmark & ViT-S/8 & COCO (118k) &  17.5  \\
    STEGO \cite{hamilton2022unsupervised} & \xmark & ViT-S/8 & COCO (118k) & 26.8 \\
    \midrule
    CLIP$\dag$ \cite{radford2021learning, zhou2022extract}  & CLIP-ResNet \hfill \: & ResNet-101 & \xmark & 6.6 \\
    MaskCLIP$\dag$ \cite{zhou2022extract}  & CLIP-ResNet \hfill \: & ResNet-101 & \xmark & 6.9 \\
    OFA$\dag$ \cite{zhou2022extract}  & OFA-Base \hfill \: & ResNet-101 & \xmark & 2.2 \\
    IFSeg (ours)$\dag$ & OFA-Base \hfill \: & ResNet-101 & \xmark & \textbf{31.0} \\
    \bottomrule
\end{tabular}
\caption{\textbf{Comparison with unsupervised semantic segmentation baselines} on the COCO Stuff benchmark. 
We report the mIoU metric evaluated on the 27 coarse semantic categories of the COCO Stuff benchmark. $\dag$ denotes that our post-processing is applied.}
\label{tbl:coco-word-augmentation}
\end{table*}

\vspace{0.02in}
\noindent
\textbf{External semantic categories from other sources.} 
Here, we validate the effect of external semantic categories from other sources. 
To this end, we perform IFSeg using the 150 semantic categories of the ADE20K benchmark~\cite{zhou2017scene} and then evaluate it on the 15 unseen categories of the COCO Stuff benchmark.\footnote{We use the same vocabulary of the unseen semantic categories of the COCO Stuff in \cref{sec:image-free}: \emph{frisbee, skateboard, cardboard, carrot, scissors, suitcase, giraffe, cow, road, wall concrete, tree, grass, river, clouds, playing field}.}
Interestingly, even though only 4 semantic categories (\emph{road}, \emph{tree}, \emph{grass}, and \emph{river}) intersect between training and evaluation, our method still achieves a significant performance of 54.1 mIoU, which is close to 54.6 mIoU of ours in \cref{tbl1:coco_unseen}, which potentially indicates that our method with external categories from other sources could learn transferable representations to novel semantic categories.

\subsection{External training images and annotations}\label{sec:external-images}
In this section, we investigate further improvements of ours when external training images and annotations as we described in \cref{sec:ablation}. Overall, we empirically found that ours can achieve the best score compared to the baselines in both \cref{tbl2:cross2ade} and \cref{tbl3:unsup_coco} when such stronger supervision is available; 
for example, ours in the last row of \cref{tbl:unsup_selftraining} shows the best score by fine-tuning task-specific images with corresponding pseudo-labels with 8k additional training iterations. 
Here, we generate pseudo-labels via our pre-trained model following Zhou \emph{et al.} \cite{zhou2022extract}.
Moreover, we also observed that fine-tuning ours in \cref{tbl2:cross2ade} with class-agnostic masks gives further enhancements from 17.4\footnote{We empirically found that using hierarchical semantic categories for the ADE20K benchmark also improves the performance from 16.8 to 17.4 mIoU score. The hierarchy is publicly available at \url{https://groups.csail.mit.edu/vision/datasets/ADE20K/}.} to 20.9 mIoU with 60k additional training iterations following the configuration of ZSSeg~\cite{xu2022simple}, which is the strongest baseline and does 20.5 on the ADE20K benchmark (see \cref{tbl:cross_ade_selftraining}). 
We note that our values using training images are reported without the post-processing, including \cref{tbl4:ablation}.

\begin{table}[h]
\centering
\small
\scalebox{0.95}{
\begin{tabular}{llcc}
    \toprule
    Method & Backbone & Image Dataset & mIoU \\
    \midrule
    IIC \cite{ji2019invariant}& ResNet-18 & COCO (118k) & 0.6 \\
    PiCIE + H. \cite{cho2021picie}& ResNet-18 & COCO (118k) & 4.6  \\
    TransFGU \cite{yin2022transfgu}& ViT-S/8 & COCO (118k) & 11.9  \\
    MaskCLIP+ \cite{zhou2022extract} & ResNet-101 & COCO (118k) & {18.0} \\
    \cmidrule{1-4}
    CLIP$\dag$ \cite{radford2021learning,zhou2022extract}& ResNet-101 & \xmark & 4.5 \\
    MaskCLIP$\dag$ \cite{zhou2022extract}& ResNet-101 & \xmark & 13.7 \\
    OFA$\dag$ \cite{wang2022ofa} & ResNet-101 & \xmark & {1.5} \\ 
    IFSeg (ours)$\dag$ & ResNet-101 & \xmark & {16.9} \\ 
    \cmidrule{1-4}
    IFSeg (ours) & ResNet-101 & COCO (118k) & \textbf{18.4} \\ 
    \bottomrule
\end{tabular}
}
\caption{\textbf{Ablation study on the effect of external training images.} All models are evaluated on the 171 semantic categories of the COCO Stuff unsupervised segmentation benchmark. The last row indicates that fine-tuned result on training images of the COCO Stuff benchmark and corresponding pseudo labels generated by ours with 8k iterations. $\dag$ denotes that our post-processing is applied.}\label{tbl:unsup_selftraining}
\end{table}
\begin{table*}[h]
\centering
\small
\scalebox{0.95}{
\begin{tabular}{lllccc}
    \toprule
    Method & Text Backbone & Image Backbone & Image Dataset & Segmentation Label & mIoU \\
    \midrule
    LSeg+ \cite{li2022languagedriven, ghiasi2022scaling} & ALIGN-BERT-Large~\cite{jia2021scaling} & ResNet-101 & COCO (118k) & \cmark & 13.0 \\
    OpenSeg \cite{ghiasi2022scaling}& ALIGN-BERT-Large~\cite{jia2021scaling} & ResNet-101 & COCO (118k) & \cmark & 15.3 \\
    ZSSeg \cite{xu2022simple} & CLIP-ViT-B~\cite{radford2021learning} & ResNet-101 & COCO (118k) & \cmark & 20.5 \\
    \cmidrule{1-6}
    CLIP$\dag$ \cite{radford2021learning,zhou2022extract} & CLIP-ResNet~\cite{radford2021learning} & ResNet-101 & \xmark & \xmark & 3.9 \\
    MaskCLIP$\dag$ \cite{zhou2022extract} & CLIP-ResNet~\cite{radford2021learning} & ResNet-101 & \xmark & \xmark & 11.3 \\
    OFA$\dag$ \cite{wang2022ofa} & OFA-Base~\cite{wang2022ofa} & ResNet-101 & \xmark & \xmark & {0.5} \\
    IFSeg (ours)$\dag$ & OFA-Base~\cite{wang2022ofa} & ResNet-101 & \xmark & \xmark & {16.8} \\
    \cmidrule{1-6}
    IFSeg (ours) & OFA-Base~\cite{wang2022ofa} & ResNet-101 & COCO (118k) & \cmark & \textbf{20.9} \\
    \bottomrule
\end{tabular}
}
\caption{\textbf{Ablation study on the effect of external segmentation annotations.} All models are evaluated on the 150 semantic categories of the ADE20K benchmark. The last row indicates that our fine-tuned result on training images of the COCO Stuff benchmark and corresponding class-agnostic segmentation masks with 60k iterations following the configuration of ZSSeg. $\dag$ denotes that our post-processing is applied.}\label{tbl:cross_ade_selftraining}
\end{table*}

\subsection{Comparison on weakly-supervised zero-shot transfer scenario}\label{supple:selftraining}
In this section, we present a comparison between our method and (weakly) supervised baselines using the self-training technique \cite{bucher2019zero}, which has been widely used in VL-driven zero-shot segmentation literature.
Inspired by the weakly-supervised zero-shot transfer recipe proposed in MaskCLIP+ \cite{zhou2022extract}, we consider a weakly-supervised variant of our model, named IFSeg+, which is trained based on the ground truth segmentation labels for the 156 seen classes, the pseudo labels for the 15 unseen classes produced by the pre-trained IFSeg,\footnote{We use the pre-trained IFSeg checkpoint having 61.6 mIoU in \cref{tbl4:ablation}.} and an additional set of pseudo labels that are produced by IFSeg+ model itself during training.
To be specific, we train IFSeg+ using the pre-trained OFA-Base \cite{wang2022ofa} checkpoint, leveraging the ground truth segmentation labels (for the seen 156 classes) and the pseudo labels (for the unseen 15 classes) generated by
the pre-trained
IFSeg during initial 15k training iterations.
Subsequently, we replace the pseudo labels generated by IFSeg with those generated by the IFSeg+ itself. We then apply the self-training technique \cite{bucher2019zero} for the remaining 66k training iterations.

For evaluation, we follow the protocol of COCO Stuff seen $\to$ unseen zero-shot transfer scenario considered by prior works \cite{bucher2019zero,gu2020context,cheng2021sign,xian2019semantic,pastore2021closer,xu2022simple,zhou2022extract} where all 171 semantic categories of the COCO Stuff have to be predicted, then the mIoU metrics for the seen and the unseen categories are individually considered (\emph{i.e.}, mIoU(U) and mIoU(S)), as well as their harmonic mean (\emph{i.e.}, hIoU).
\cref{tbl:ablation-seen-to-unseen} summarizes the results; our method (\emph{i.e.}, IFSeg+) can achieve significant segmentation performances compared to all the baselines. For example, IFSeg+ scored 2.1, 3.7, and 3.2 higher points than MaskCLIP+ \cite{zhou2022extract} in terms of mIoU(U), mIoU(S), and hIoU, respectively.
We note that our post-processing technique is not applied to the weakly-supervised zero-shot models, as the effect of the technique diminishes after using the real images and annotations during training as discussed in \cref{sec2:ifseg}. 
For example, applying the post-processing ($K=3$ with 25 iterations)
even degrades the mIoU(U) scores of IFSeg+ and MaskCLIP+, dropping from 56.8 to 55.2, and from 54.7 to 54.5, respectively.

\begin{table*}[h]
\centering
\small
\scalebox{0.92}
{
\begin{tabular}{lllccccc}
    \toprule
    Method & Text Backbone & Image Backbone & Image Dataset & 
Segmentation Label & mIoU(U) & mIoU(S) & hIoU \\
    \midrule
    ZS5 \cite{bucher2019zero} & word2vec~\cite{mikolov2013distributed} & ResNet-101 & COCO (118k) & \cmark (156 seen)  & 10.6 & 34.9 & 16.2 \\
    CaGNet \cite{gu2020context} & word2vec~\cite{mikolov2013distributed}, fasttext~\cite{joulin2016fasttext} & ResNet-101 & COCO (118k) & \cmark (156 seen) & 13.4 & 35.3 & 32.6 \\
    SIGN \cite{cheng2021sign} & word2vec~\cite{mikolov2013distributed}, fasttext~\cite{joulin2016fasttext} & ResNet-101 & COCO (118k) & \cmark (156 seen) & 15.2 & 36.4 & 21.4 \\
    SPNet \cite{xian2019semantic} & word2vec~\cite{mikolov2013distributed}, fasttext~\cite{joulin2016fasttext} & ResNet-101 & COCO (118k) & \cmark (156 seen) & 26.9 & 34.6 & 30.3 \\
    STRICT \cite{pastore2021closer} & word2vec~\cite{mikolov2013distributed}, fasttext~\cite{joulin2016fasttext} & ResNet-101 & COCO (118k) & \cmark (156 seen) & 30.3 & 35.3 & 32.6 \\
    ZSSeg \cite{xu2022simple}  & ALIGN-BERT-Large~\cite{jia2021scaling} & ResNet-101 & COCO (118k) & \cmark (156 seen) & 43.6 & 39.6 & 41.5 \\
    MaskCLIP+ \cite{zhou2022extract} & CLIP-ResNet \cite{radford2021learning} & ResNet-101 & COCO (118k) & \cmark (156 seen) & 54.7 & 38.2 & 45.0 \\
    IFSeg+ (ours) & OFA-Base \cite{radford2021learning} & ResNet-101 & COCO (118k) & \cmark (156 seen) & \textbf{56.8} & \textbf{41.9} & \textbf{48.2} \\
    \bottomrule
\end{tabular}
}
\caption{\textbf{Comparison with (weakly) supervised baselines under the seen$\rightarrow$unseen transfer scenario.} We report the mIoU metric evaluated on the 15 unseen and the 156 seen semantic categories of the COCO Stuff benchmark and their harmonic mean, denoted by mIoU(U), mIoU(S), and hIoU, respectively.  All models are trained on segmentation labels of the 156 seen categories (supervised training) and pseudo-labels of the 15 unseen categories (self-training), where ``Image Dataset'' denotes the dataset required for training.}
\label{tbl:ablation-seen-to-unseen}
\end{table*}

\section{Ablation study on hyperparameters}
\label{supple:abl:hyperparameter}
In this section, we perform an ablation study to understand the effect of hyperparameters of our method, namely the iteration count and $K$-nearest neighbors used in the post-processing, the sampling range $S$ for the artificial image, and the use of cross-attention mechanism in our transformer decoder.

\vspace{0.02in}
\noindent
\textbf{Post processing.}
We first examine the effect of the iteration count and the number of nearest neighbor $K$ in our post-processing across an array of $\{0, 1, 10, 25, 50\}$ iteration count and $K\in\{2, 3, 5, 8\}$. As shown in \cref{tbl:ablation-postprocessing}, the effect of iteration counts becomes saturated after 25 iterations, and our method could be further improved with a larger $K$ (\emph{e.g.}, $K=8$). We note that the evaluations are performed under the zero-shot semantic segmentation on the 15 unseen semantic categories of the COCO Stuff. We also note that 0 iteration is equivalent to not performing the post-processing.

\begin{table*}[h]
\begin{subfigure}{0.5\linewidth}
\centering
    \begin{tabular}{lccccc}
	\toprule
	Iteration
	& 0 & 1 & 10 & 25 & 50\\\midrule
mIoU& 46.8 & 51.2 & 54.9 & 55.6 & 55.5 \\ 
	\bottomrule
    \end{tabular}
    \caption{Varying iteration counts with $K=3$.}
    \label{supp:detect}
\end{subfigure}
\begin{subfigure}{0.5\linewidth}
\centering
\begin{tabular}{lcccc}
    \toprule
        $K$     & 2 & 3 & 5 & 8  \\\midrule
    mIoU & 50.1 & 55.6 & 59.7 & 61.4 \\
        \bottomrule
\end{tabular}
    \caption{Varying $K$ with iteration counts of 25.}
\end{subfigure}
\caption{\textbf{Ablation studies on varying the iteration count and the number of nearest neighbor $K$.} All models are trained and evaluated on the 15 unseen semantic categories of the COCO Stuff benchmark.}\label{tbl:ablation-postprocessing}
\end{table*}

\begin{table}[h]
\centering
\small
\scalebox{0.95}{
\begin{tabular}{lclccc}
    \toprule
    Method & PP & Backbone & Zero-shot (mIoU) & Cross-dataset (mIoU) & Unsupervised (mIoU) \\
    \midrule
    CLIP \cite{radford2021learning,zhou2022extract} & \cmark & ResNet-101 & \textbf{12.3} & \textbf{3.7} & \textbf{4.6} \\
    CLIP \cite{radford2021learning,zhou2022extract} & \xmark & ResNet-101 & 11.6 & 3.6 & 4.4 \\
    \midrule
    MaskCLIP \cite{zhou2022extract} & \cmark & ResNet-101 & \textbf{24.8} & \textbf{10.3} & \textbf{12.7} \\
    MaskCLIP \cite{zhou2022extract} & \xmark & ResNet-101 & 23.7 & 8.8 & 10.8 \\
    \midrule
    IFSeg (ours) & \cmark & ResNet-101 & \textbf{54.6} & \textbf{16.8} & \textbf{16.9} \\
    IFSeg (ours) & \xmark & ResNet-101 & {47.0} & {14.0} & {14.3} \\
    \bottomrule
\end{tabular}
}
\caption{\textbf{Effects of the post-processing on varying image-free approaches.} We report the mIoU metric with and without the post-processing, evaluated on the zero-shot (the 15 unseen categories in COCO Stuff), the cross-dataset (COCO$\rightarrow$ADE20K), and the unsupervised (all the 171 categories in COCO Stuff) semantic segmentation scenarios. ``PP'' denotes our post-processing is applied.}
\label{tbl:ablation-basline-without-postprocessing}
\end{table}

Next, we present the mIoU of the image-free models (\emph{i.e.}, CLIP\cite{radford2021learning, zhou2022extract}, MaskCLIP\cite{zhou2022extract}, and IFSeg) without our post-processing evaluated on the zero-shot (the 15 unseen categories in COCO Stuff), the cross-dataset (COCO$\rightarrow$ADE20K), and the unsupervised semantic segmentation (all the 171 categories in the COCO Stuff) scenarios in \cref{tbl:ablation-basline-without-postprocessing}. 
Overall, our post-processing positively affects the mIoU of all baselines (\emph{e.g.}, 23.7 mIoU $\to$ 24.8 mIoU for MaskCLIP on the zero-shot semantic segmentation scenario). Regardless of whether or not the post-processing is applied, however, IFSeg is always the best-performing image-free model in all the scenarios.

\vspace{0.02in}
\noindent
\textbf{Artificial image.}
Here, we investigate the effect of varying sampling range $k$ for our artificial image generation.
\cref{tbl:ablation-artificial-image-k} summarizes results; interestingly, optimal values of $k=16$ and $K=8$ (of the post-processing) give our significant further improvements from 55.6 to 66.0 (+ 10.4) mIoU score on the 15 unseen semantic categories of the COCO Stuff benchmark. 
We remark that the values of $h$ and $w$ in \cref{eq:artificial_image_hw} are randomly sampled from $\{1, 2, ..., k\}$.
Regarding this, the last row in \cref{tbl:ablation-artificial-image-k} shows that removing randomness from sampling $h$ and $w$ harms overall improvements.

\begin{table}[h!]
\centering
\begin{tabular}{c l c c}
    \toprule
$S$&    $(h, w) \sim \{1, 2, ..., S\}$ & Post-processing with $K=3$ & Post-processing with $K=8$\\
    \midrule
8&    $(h, w) \sim \{1, 2, ..., 8\}$ & 55.8 & 64.3 \\
16&    $(h, w) \sim \{1, 2, ..., 16\}$ & \textbf{57.8} & \textbf{66.0}\\
32&    $(h, w) \sim \{1, 2, ..., 32\}$ & 55.6 & 61.4 \\\cmidrule{1-4}
32&    $(h, w) = (32, 32)$ & 47.7 & 56.1 \\
    \bottomrule
\end{tabular}
\caption{\textbf{Ablation studies on varying sampling range $S$ for our artificial image generation.} We also report two different nearest neighbor hyperparameters $K\in\{3,8\}$ of the post-processing. The last row reports the deterministic setup of $(h,w)=(32,32)$ for generating our artificial images. The reported values are mIoU scores on the 15 unseen semantic categories of the COCO Stuff benchmark.}
\label{tbl:ablation-artificial-image-k}
\end{table}
On the other hand, one may consider the recent VL prompt learning method \cite{zhou2022learning, zhou2022conditional} as an option for efficiently adapting a VL model to the semantic segmentation task. However, we would like to emphasize that our primary interest lies in image-free scenarios. Simply plugging the prompt learning into the image-free setting is non-trivial, as prompting cannot replace the training images and labels required to learn the segmentation task. 
Nonetheless, formulating image-free semantic segmentation within the context of the prompt learning framework could be an interesting direction for future research.

\vspace{0.02in}
\noindent \textbf{The cross-attention mechanism.}
We validate the effect of the cross-attention mechanism in our transformer decoder.
To this end, we train our model on the zero-shot (the 15 unseen categories in COCO Stuff) semantic segmentation scenario without providing the contextualized embedding (\cref{eq:transformer_encoder}) for the cross-attention mechanism.
As a result, we observed a significant degradation in segmentation performance, dropping from 55.6 $\rightarrow$ 22.6 mIoU after removing the cross-attention. 
We note that the use of cross-attention is a default setting during the VL pre-training in our framework, and maintaining the cross-attention during fine-tuning would be beneficial for stability.

\section{Image-free baselines with ViT backbone}
In this section, we present the mIoU of the image-free baselines, CLIP\cite{radford2021learning,zhou2022extract} and MaskCLIP\cite{zhou2022extract}, with the stronger ViT-B/16 image backbones evaluated on the zero-shot (the 15 unseen categories in COCO Stuff) semantic segmentation scenario in \cref{tbl:ablation-vit-backbone}. Overall, ViT-B/16 brings performance improvements thanks to its advanced visual representation compared to the ResNet-101 backbone. Nonetheless, the performance of our IFSeg is superior to these baselines even if it uses the ResNet-101 as the image backbone model, unchanged from the trends observed in \cref{tbl1:coco_unseen}.

\begin{table}[h]
\centering
\small
\scalebox{0.95}{
\begin{tabular}{lllc}
    \toprule
    Method & Text Backbone & Image Backbone & mIoU \\
    \midrule
    CLIP$\dag$ \cite{radford2021learning,zhou2022extract} & CLIP-ViT-B/16~\cite{radford2021learning} & ViT-B/16 & 12.9 \\
    MaskCLIP$\dag$ \cite{zhou2022extract} & CLIP-ViT-B/16~\cite{radford2021learning}  & ViT-B/16 & 37.0 \\
    IFSeg (ours)$\dag$ & OFA-Base~\cite{wang2022ofa} & ResNet-101  & \textbf{55.6} \\
    \bottomrule
\end{tabular}
}
\caption{\textbf{Comparison with image-free baselines under the zero-shot semantic segmentation (the 15 unseen categories in COCO Stuff) scenario.} We report the mIoU metric evaluated on the 15 unseen semantic categories of the COCO Stuff benchmark. $\dag$ indicates models with our post-processing applied.}
\label{tbl:ablation-vit-backbone}
\end{table}

\section{Compatibility analysis}
We here validate the compatibility of our method with another encoder-decoder VL model, CLIPCap~\cite{mokady2021clipcap}. 
Note that CLIPCap is a fine-tuned CLIP-ViT-B/32 model for an image-to-text captioning task on the Conceptual Captions benchmark~\cite{sharma2018conceptual}. Specifically, CLIPCap utilizes GPT2~\cite{radford2019language} as a text generator, and we also do it as the segmentation decoder in our framework. 

In order to create our artificial image under CLIP's dual-encoder design, we utilize CLIP text encoder's sentence-level feature as the word embedding for semantic categories, directly following the prompt engineering procedure by MaskCLIP \cite{zhou2022extract}. For example, an artificial image patch for a \emph{dog} category is an ensemble of prompts like ``\emph{a photo of the dog}'' and ``\emph{a painting of a dog}.''\footnote{We refer the readers to the codebase of MaskCLIP \cite{zhou2022extract} for the full list of prompt templates; \url{https://github.com/chongzhou96/MaskCLIP}.} 
Then, similar to ours incorporated with the OFA framework, we fine-tune the text generator of ClipCap to predict semantic segmentation of the artificial image and evaluate the performance on the 15 unseen semantic categories of the COCO Stuff benchmark.
We note that, in order to deal with the prefix-based design of ClipCap (\emph{i.e.}, a single token in the CLIP representation space is mapped to multiple tokens in the text generator space), we decode each token individually.

\cref{tbl:clipcap-results} summarizes the compatibility experiments; our method is well-incorporated with CLIPCap and even significantly outperforms CLIP and MaskCLIP~\cite{zhou2022extract} baselines, which also have the same CLIP backbone. 
For example, our method achieves the best mIoU score of 25.8 on the 15 unseen semantic categories of the COCO Stuff benchmark compared to the baselines having the same CLIP backbone.
These results demonstrate the broad applicability of our method with various pre-trained VL models and lead them to perform semantic segmentation in an image-free manner.

\begin{table}[h]
\centering
\small
\begin{tabular}{lllcc}
    \toprule
    Method & Pretrain & Image Backbone & Text Deocder & mIoU \\
    \midrule
    CLIP$\dag$ \cite{zhou2022extract, radford2021learning}  & CLIP \cite{radford2021learning} & CLIP-ViT-B/32 & \xmark & 4.8 \\
    MaskCLIP$\dag$ \cite{zhou2022extract}  & CLIP \cite{radford2021learning} & CLIP-ViT-B/32 & \xmark & 20.7 \\
    IFSeg (ours)  & CLIPCap \cite{mokady2021clipcap} & CLIP-ViT-B/32 & GPT2~\cite{radford2019language} & \textbf{25.8} \\
    \bottomrule
\end{tabular}
\caption{\textbf{Ablation study on compatibility with other encoder-decoder VL models.} We denote that our image-free approach is applied to CLIPCap, which is an image captioning model built upon pre-trained CLIP. 
All models are evaluated on the 15 unseen semantic categories of the COCO Stuff benchmark. $\dag$ denotes that our post-processing is applied.}
\label{tbl:clipcap-results}
\end{table}

\section{Implementation details}
\vspace{0.02in}
\noindent \textbf{Image Pre-processing.} We preprocess images using the official codebase\footnote{\url{https://github.com/OFA-Sys/OFA}.} of OFA \cite{wang2022ofa} framework and $\tt {mmsegmentation}$\footnote{\url{https://github.com/open-mmlab/mmsegmentation}.}. Specifically, we normalize the image with the mean and standard deviation values of 0.5. We also resize the short sides of images keeping the aspect ratio. 
For all experiments, we resize the short sides to 512, in order to ensure a fair comparison with the strongest baselines MaskCLIP+~\cite{zhou2022extract} and DenseCLIP~\cite{rao2022denseclip}.

\vspace{0.02in}
\noindent \textbf{Text Pre-processing.} We generate prompt text following the ``\textit{task description} $+$ \textit{category enumeration}'' protocol of the VQA task~\cite{wang2022ofa}. Precisely, we use ``\textit{what is the segmentation of the image?}'' as the task description, and ``\textit{object: category1 category2 ... categoryN}'' as the category enumeration. For the tokenization and embedding, we directly incorporate the pre-trained BPE tokenizer and embedding matrix provided by the codebase of OFA \cite{wang2022ofa} framework.

\vspace{0.02in}
\noindent \textbf{Evaluation Details.} For a fair comparison, we perform the \emph{whole inference} evaluation protocol (\emph{i.e.}, predicting the rectangular-shaped output at once) for image-free based approaches (\emph{e.g.}, \cref{tbl1:coco_unseen,tbl2:cross2ade,tbl3:unsup_coco,tbl4:ablation}) following their strongest baseline, MaskCLIP+~\cite{zhou2022extract}, and the \emph{sliding inference} evaluation protocol (\emph{i.e.}, concatenating square-shaped crops of the original rectangular-shaped image) for supervised approaches (\emph{e.g.}, \cref{tbl5:supervised_ade}) following their strongest baseline, DenseCLIP \cite{rao2022denseclip}. 

\vspace{0.02in}
\noindent \textbf{Visual Feature-based Post-processing.} 
In our post-processing,
we utilize features of the image backbone network (\emph{i.e.}, ResNet) from the OFA \cite{wang2022ofa} encoder.
For a fair comparison, we also apply the post-processing for our re-implemented baselines of OFA \cite{wang2022ofa}, CLIP \cite{radford2021learning}, and MaskCLIP \cite{zhou2022extract} with their image backbone networks. 
For example, in the case of CLIP \cite{radford2021learning} and MaskCLIP \cite{zhou2022extract}), we utilize the final patch-wise outputs of the ViT-B/16 image backbone as the post-processing features.

\vspace{0.02in}
\noindent \textbf{Viusalization Details.} Exclusively for the visualizations of the image-free models (\cref{fig1:first-page,fig4:visualize_ifseg}), we introduce additional post-processing with DenseCRF \cite{krahenbuhl2011efficient} and its third-party implementation\footnote{\url{https://github.com/lucasb-eyer/pydensecrf}.}. Note that smoothing outputs with DenseCRF can provide qualitatively sharper segmentation results by clustering prediction outputs according to the edges of the raw RGB images. 
However, we remark that DenseCRF is \emph{never} used for the reported values of experimental results for a fair comparison.

\begin{table*}[h!]
\centering
\scalebox{0.9}{
\begin{tabular}{ll}
    \toprule
    Coarse category & Fine-grained category \\
    \midrule
    animal & giraffe, zebra, bear, elephant, cow, sheep, horse, dog, cat, bird \\
    sports & tennis racket, surfboard, skateboard, baseball glove, baseball bat, kite, sports ball, snowboard, skits, frisbee \\
    accessory & suitcase, tie, handbag, eye glasses, shoe, umbrella, backpack, hat \\
    outdoor & bench, parking meter, stop sign, street sign, fire hydrant, traffic light \\
    vehicle & boat, truck, train, bus, airplane, motorcycle, car, bicycle \\
    person & man, woman, child, boy, girl \\
    indoor & hair brush, toothbrush, hair drier, teddy bear, scissors, vase, clock, book \\
    appliance & blender, refrigerator, sink, toaster, oven, microwave \\
    electronic & cell phone, keyboard, remote, mouse, laptop, tv \\
    furniture (things) & door, toilet, desk, window, dining table, mirror, bed, potted plant, couch, chair \\
    food (things) & cake, donut, pizza, hot dog, carrot, broccoli, orange, sandwich, apple, banana \\
    kitchen & bowl, spoon, knife, fork, cup, wine glass, plate, bottle \\
    water & waterdrops, sea, river, fog, lake, ocean \\
    ground & playingfield, platform, railroad, pavement, road, gravel, mud, dirt, snow, sand \\
    solid & hill, mountain, stone, rock, wood \\
    sky & clouds \\
    plant & straw, moss, branch, flower, leaves, bush, tree, grass \\
    structural & railing, net, cage, fence \\
    building & roof, tent, bridge, skyscrapper, house \\
    food (stuff) & vegetable, salad, fruit \\
    textile & banner, pillow, blanket, curtain, cloth, clothes, napkin, towel, mat, rug \\
    furniture (stuff) & stairs, light, counter, mirror, cupboard, cabinet, shelf, table, desk, door \\
    window & blind window \\
    floor & stone floor, marble floor, wood floor, tile floor, carpet \\
    ceiling & tile ceiling \\
    wall & concrete wall, stone wall, brick wall, wood wall, panel wall, tile wall \\
    raw material & metal, plastic, paper, cardboard \\
    \bottomrule
\end{tabular}
}
\caption{\textbf{The full list of hierarchical semantic categories} of the COCO Stuff benchmark. Each coarse category is paired with given fine-grained categories, following the label hierarchy of Caesar \emph{et al.}~\cite{caesar2018coco}.}
\label{tbl:coco-coarse-mapping}
\end{table*}

\section{Additional qualitative results}
In this section, we present visualizations of segmentation results obtained by baselines and our method in different evaluation settings. Specifically, we first consider the weakly-supervised scenario (zero-shot transfer) in \cref{tbl:ablation-seen-to-unseen} and compare the result between our IFSeg+ and MaskCLIP+, the strongest baseline in the scenario. Next, we also consider the fully-supervised semantic segmentation scenario in \cref{tbl5:supervised_ade} and compare the result between the supervised IFSeg and DenseCLIP baseline.

\subsection{Weakly-supervised zero-shot transfer scenario}
\begin{figure*}[h!]
\vspace{-0.1in}
 \centering
 \scalebox{1.0}{
  \includegraphics[width=0.93\textwidth]{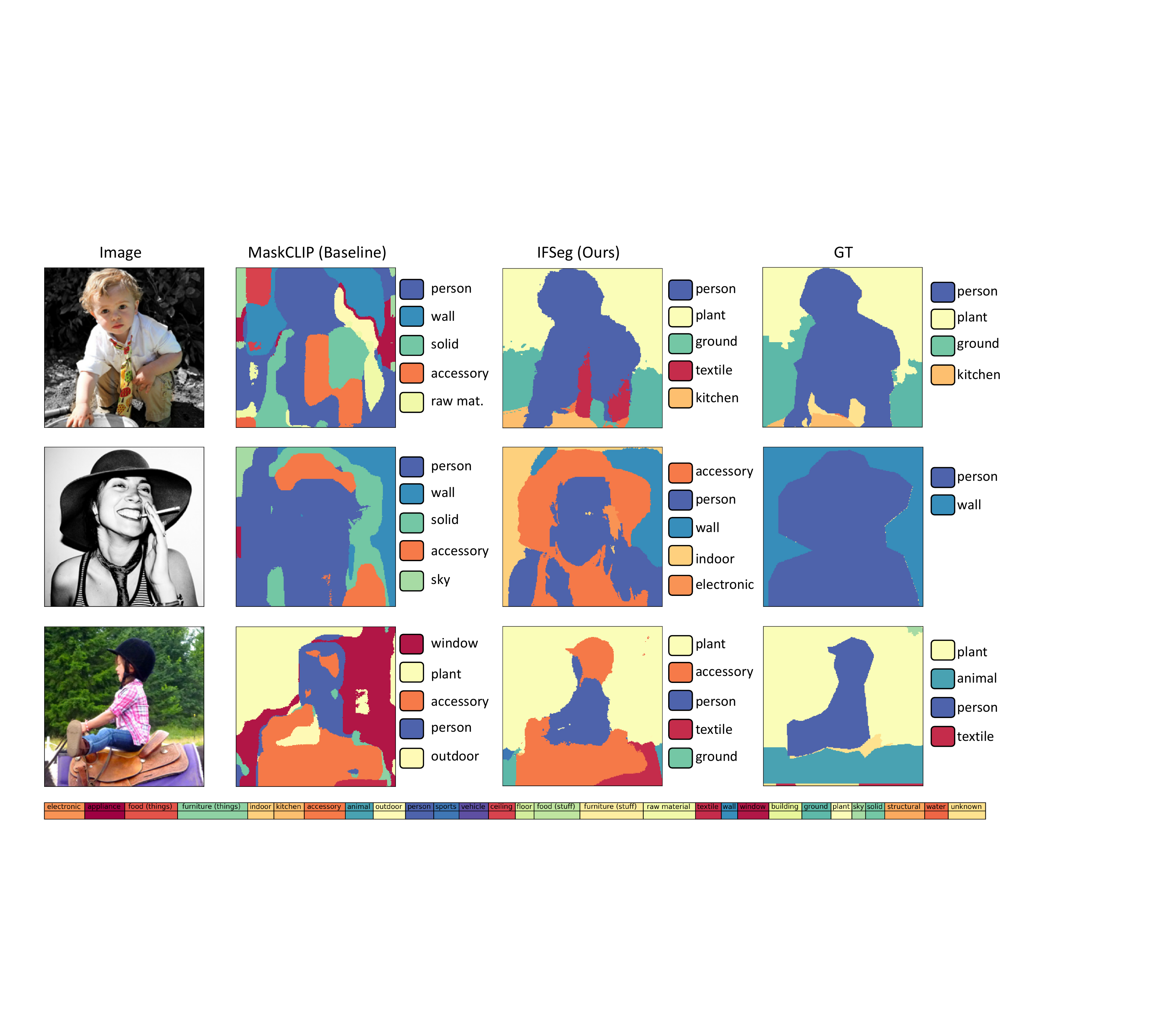}
  }
\caption{\textbf{Visualization of segmentation results under the weakly-supervised zero-shot transfer scenario.} We visualize the segmentation results of IFSeg+ (ours) and MaskCLIP+ (baseline). Qualitatively observed, IFSeg+ can predict the segmentation that is more consistent with the groud-truth (GT) segmentation than the MaskCLIP+ baseline. Best viewed in color.}
\label{fig:visualization-coco-unseen}
\vspace{-0.1in}
\end{figure*}

The visualizations of segmentation results obtained by MaskCLIP+ and ours under the COCO Stuff seen$\to$unseen zero-shot transfer scenario are present in \cref{fig:visualization-coco-unseen}. Following the protocol in \cref{supple:selftraining} we evaluate and visualize the 15 unseen classes of the COCO Stuff benchmark. Overall, it shows that our method can predict the segmentation that is more consistent with the groud-truth (GT) segmentation than the MaskCLIP+ baseline.

\subsection{Fully-supervised semantic segmentation scenario}
\begin{figure*}[h!]
\vspace{-0.1in}
 \centering
 \scalebox{0.97}{
  \includegraphics[width=0.93\textwidth]{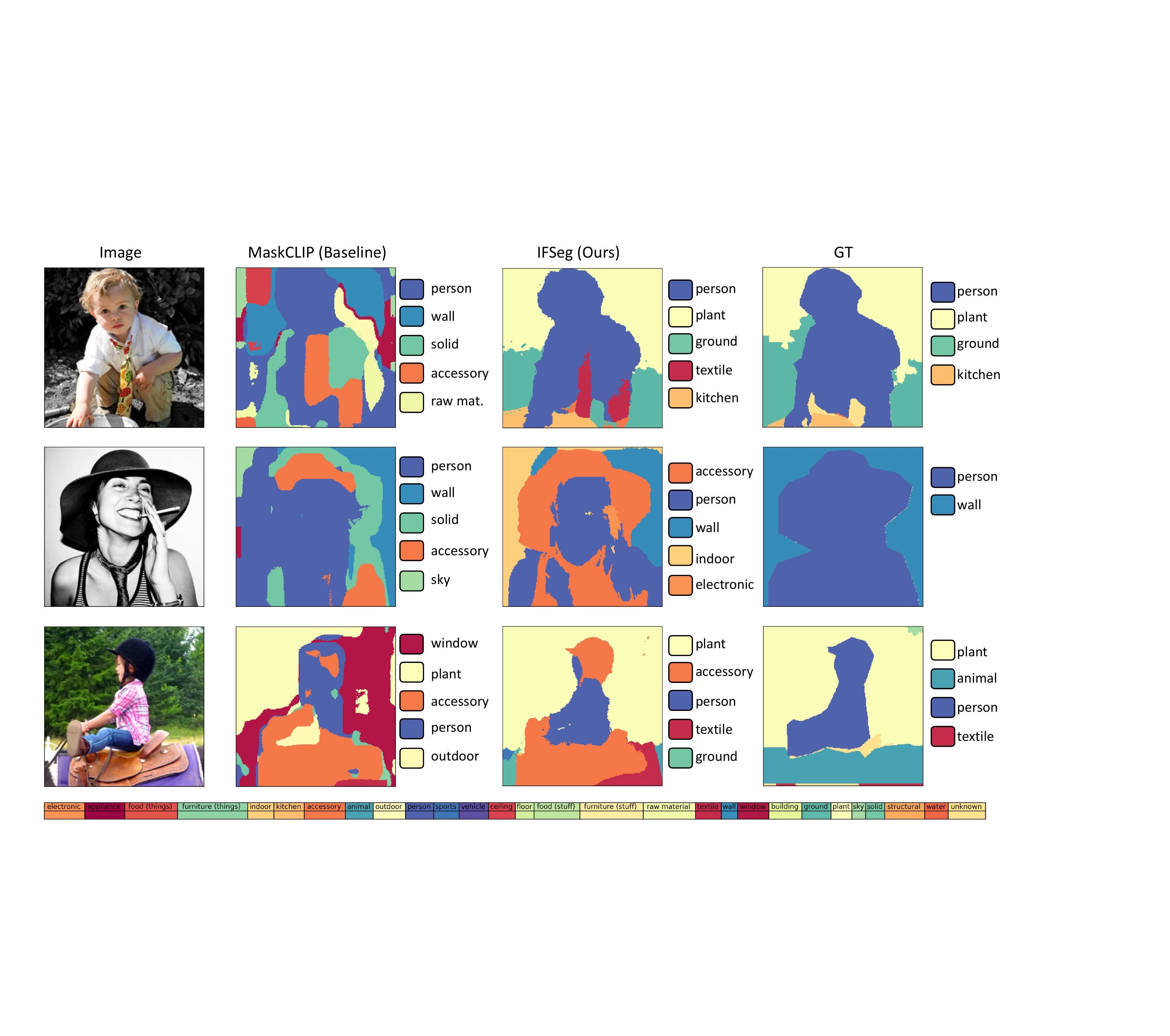}
  }
\caption{{\textbf{Visualization of segmentation results under the fully-supervised semantic segmentation scenario.} We visualize the segmentation results of Supervised IFSeg (ours) and DenseCLIP (baseline). Although both the models are trained in a fully-supervised manner, our IFSeg tends to produce more accurate predictions than DenseCLIP. We utilize the class colors defined by the $\tt {mmsegmentation}$ in \url{https://github.com/open-mmlab/mmsegmentation/blob/master/mmseg/datasets/ade.py}. For clarity, we denote the 14 classes with the largest segmentation regions in this example.}
Best viewed in color.}
\label{fig:visualization-ade}
\end{figure*}

The visualizations of segmentation results obtained by DenseCLIP and ours under the ADE20k semantic segmentation benchmark are present in \cref{fig:visualization-ade}. We evaluate and visualize the total 150 classes of the ADE20k dataset. As depicted by the quantitative mIoU score in \cref{tbl5:supervised_ade} and some visualization cases in \cref{fig:visualization-ade}, ours shows results that are more consistent with the groud-truth (GT) segmentation than the DenseCLIP baseline. However, we note that DenseCLIP and ours both tend to produce satisfactory prediction results for most samples since they are trained in a fully-supervised way using the ground-truth segmentation annotations.

\end{document}